\documentclass{article}

\usepackage[nonatbib,preprint]{neurips_2020}
\usepackage[numbers]{natbib}

\usepackage[utf8]{inputenc} 
\usepackage[T1]{fontenc}    
\usepackage{hyperref}       
\usepackage{url}            
\usepackage{booktabs}       
\usepackage{amsfonts}       
\usepackage{nicefrac}       
\usepackage{microtype}      

\usepackage{graphicx}
\usepackage{subfigure}
\usepackage{booktabs} 

\usepackage{amsmath,amsfonts,bm}

\def\eqref#1{equation~\ref{#1}}

\def\1{\bm{1}}

\DeclareMathAlphabet{\mathsfit}{\encodingdefault}{\sfdefault}{m}{sl}
\SetMathAlphabet{\mathsfit}{bold}{\encodingdefault}{\sfdefault}{bx}{n}

\newcommand{\sigmoid}{\sigma}

\usepackage{algorithm}
\usepackage{amssymb}
\usepackage{mathtools}
\usepackage{cancel}
\usepackage{booktabs}
\usepackage{comment}
\usepackage{tikz}
\usepackage{asypictureB}
\usepackage{bm}
\usepackage{caption}
\usepackage{wrapfig}

\usepackage[utf8]{inputenc} 
\usepackage[T1]{fontenc}    
\usepackage{hyperref}       
\usepackage{url}            
\usepackage{booktabs}       
\usepackage{amsfonts}       
\usepackage{nicefrac}       
\usepackage{microtype}      
\usepackage[utf8]{inputenc} 
\usepackage[T1]{fontenc}    
\usepackage{hyperref}       
\usepackage{url}            
\usepackage{amsfonts}       
\usepackage{nicefrac}       
\usepackage{graphicx}
\usepackage{booktabs,multirow} 
\usepackage{amsmath}
\usepackage{makecell}
\usepackage[export]{adjustbox}

\usepackage{verbatim} 

\usepackage{amssymb}
\usepackage{mathtools}
\usepackage{cancel}
\usepackage{booktabs}
\usepackage{comment}
\usepackage{tikz}
\usepackage{asypictureB}
\usepackage{bm}
\usepackage{caption}
\usepackage{wrapfig}
\usetikzlibrary{backgrounds}
\usetikzlibrary{calc}
\usetikzlibrary{fit}
\usetikzlibrary{positioning}
\usepackage{enumitem}
\usepackage[acronym,smallcaps,nowarn]{glossaries}
\usepackage{soul}
\usepackage{hyperref}
\usepackage[font={scriptsize}]{caption}
\usepackage{todonotes}
\usepackage{multicol}
\usepackage[noend]{algpseudocode}
\usepackage{float,pbox}
\usepackage{pgfplots}
\pgfplotsset{compat=1.14}
\usepackage{setspace}
\usepackage{todonotes}
\usepackage[toc,page]{appendix}

\newcommand\mc{\multicolumn}

\newcommand{\highlight}[1]{\colorbox{blue!10}{#1}}

\newacronym{SDI}{sdi}{Structural Discovery from Interventions}
\newacronym{NIMs}{nims}{Neural Interventional Models}
\newacronym{NIM}{nim}{Neural Interventional Model}

\title{Learning Neural Causal Models from Unknown Interventions}

\author{
 Nan Rosemary Ke$^\textbf{*}$\,$^{1,2}$,
Olexa Bilaniuk$^\textbf{*}$\,$^{1}$,
Anirudh Goyal$^{1}$, 
Stefan Bauer$^{5}$, \\
{\bf Hugo Larochelle$^{4}$}, 
{\bf Bernhard Schölkopf$^{5}$},
{\bf Michael C. Mozer$^{4}$},
{\bf Chris Pal$^{1,2,3}$},
{\bf Yoshua Bengio$^{1\dagger}$}\\
\\
$^1$ Mila, Universit\'e de Montr\'eal,
$^2$ Mila, Polytechnique Montr\'eal,
$^3$ Element AI \\
$^4$ Google AI,
$^5$ Max Planck Institute for Intelligent Systems,
$^\dagger$CIFAR Senior Fellow.\\
$^\textbf{*}$ \textbf{Authors contributed equally},
\texttt{rosemary.nan.ke@gmail.com}
}

\begin{document}

\maketitle

\begin{abstract}

Promising results have driven a recent surge of interest in continuous optimization methods for Bayesian network structure learning from observational data. However, there are theoretical limitations on the identifiability of underlying structures obtained from observational data alone. Interventional data provides much richer information about the underlying data-generating process. However, the extension and application of methods designed for observational data to include interventions is not straightforward and remains an open problem. In this paper we provide a general framework based on continuous optimization and neural networks to create models for the combination of observational and interventional data. The proposed method is even applicable in the challenging and realistic case that the identity of the intervened upon variable is unknown. We examine the proposed method in the setting of graph recovery both de novo and from a partially-known edge set. We establish strong benchmark results on several structure learning tasks, including structure recovery of both synthetic graphs as well as standard graphs from the Bayesian Network Repository.
\end{abstract}

\section{Introduction}
\iftrue

Structure learning concerns itself with the recovery of the graph structure of Bayesian networks from data. When Bayesian networks are used to model cause-effect relationships and are augmented with the notion of \textit{interventions} and counterfactuals, they can be represented as structural causal models (SCM).
While Bayesian networks can uncover statistical correlations between factors, SCMs can be used to answer higher-order questions of cause-and-effect, up in the ladder of causation \citep{pearl2018book}. Causal structure learning using SCMs has been attempted in several disciplines including biology \citep{sachs2005causal,hill2016inferring}, weather forecasting \citep{abramson1996hailfinder} and medicine \citep{lauritzen1988local,korb2010bayesian}.

Structure can be learned from data samples drawn from observational or interventional distributions. The data can be more or less revelatory about the underlying structure, which affects a structure learning method's ability to identify structure using that data. Observational data is sampled from the distribution without interventions; alone, it contains only limited information about the underlying causal graph and hence structure learning methods generally cannot do more than identify the causal graph up to a Markov equivalence class \citep{spirtes2000causation}. In order to fully identify the true causal graph, a method must either make restrictive assumptions about the underlying data-generating process, such as linear but non-Gaussian data \citep{shimizu2006linear}, or must access enough data from outside the observational distribution (i.e., from interventions). Under certain assumptions about the number, diversity, and nature of the interventions, the true underlying causal graph is always identifiable, given that the method knows the intervention performed \citep{heckerman1995learning}. However, in the real world, interventions can often be performed by other agents, which would make them \textit{unknown interventions}. A few works have attempted to learn structures from unknown-intervention data \citep{eaton2007belief,mooij2016joint,tillman2011learning,rothenhausler2015backshift}. Although there is no theoretical guarantee that the true causal graph can be identified in that setting, evidence so far points to that still being the case.

Another common setting is when the graph structure is partially provided, but must be completed. An example is protein structure learning in biology, where we may have definite knowledge about some parts of the protein-protein interaction structure and have to fill out other parts. We will call this setting partial graph completion. This is an easier task compared to learning the entire graph, since it limits the number of edges that have to be learned.

\begin{wrapfigure}{r}{0.5\textwidth}
    \includegraphics[width=.5\textwidth]{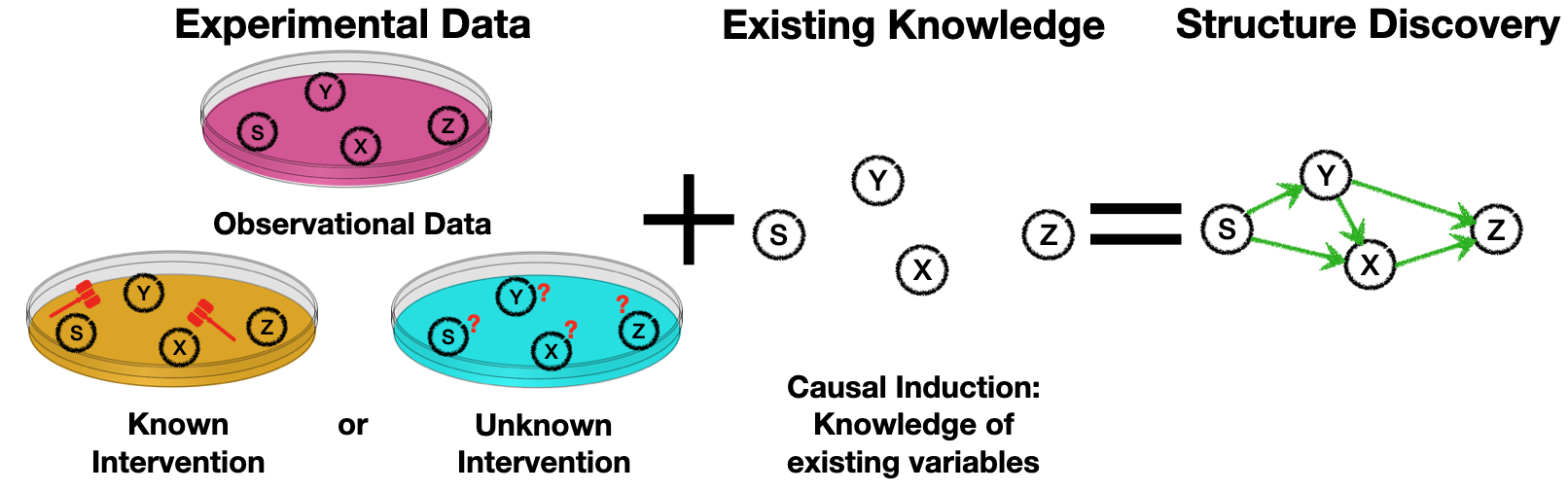}
    \caption{In many areas of science, such as  biology, we try to infer the underlying mechanisms and structure through experiments. We can obtain observational data plus interventional data through known (e.g. by targeting a certain variable) or unknown interventions (e.g. when it is unclear where the effect of the intervention will be). Knowledge of existing edges e.g. through previous experiments can likewise be included and be considered a special case of causal induction. 
    }
    \vspace{-1\baselineskip}
    \label{fig:problem_statement}
\end{wrapfigure}

Recently, a flurry of work on structure learning using continuous optimization methods has appeared \citep{zheng2018dags,yu2019dag}. These methods operate on observational data and perform competitively to other methods. There are theoretical limitations on how well the causal graph can be identified from only observational data and it would be interesting to apply these methods on interventional data, which offers more information about the underlying causal structure. However, it is not straightforward to apply continuous optimization methods to structure learning from interventional data. 
\textbf{Contributions}, we experimentally answer the following questions:
\vspace{-1mm}
\begin{enumerate}
\item Can the proposed model recover true causal structure? Yes, see Figure \S\ref{fig:synthetic_data_results}.
\item How does the proposed model compare against state of the art causal methods on real-world datasets? \textit{Favourably;} see \S\ref{baseline_compare} and Table \S\ref{table:all_baseline_hamming}.
\item Does a proposed model generalize well to unseen interventions? Yes, see \S\ref{generalization}.
\item How does the proposed model perform on partial graph recovery? It scales to $\sim 50$ variables; see \S\ref{partial)_graph}.
\end{enumerate}
\section{Preliminaries}
\paragraph{Causal modeling.}A Structural Causal Model (SCM) \citep{peters2017elements}  over a finite number $M$ of random variables $X_i$ is a set of structural assignments
\begin{align}\label{eq:structuralassignment}
	X_i &:= f_i(X_{pa(i,C)}, N_i)\;, \quad \forall i \in \{0,\ldots,M-1\}
\end{align}
where $N_i$ is jointly-independent noise and $pa(i,C)$ is the set of parents (direct causes) of variable $i$ under hypothesized configuration $C$ of the SCM directed acyclic graph, i.e., $C \in \{0,1\}^{M \times M}$, with $c_{ij} = 1$ if node $i$ has node $j$ as a parent (equivalently, $X_j \in X_{pa(i,C)}$; i.e. $X_j$ is a direct cause of $X_i$). 

\paragraph{Identifiability.} 
In a purely-observational setting, it is known that causal graphs can be distinguished only up to a Markov equivalence class. In order to identify the true causal graph structure, intervention data is needed \citep{eberhardt2012number}.
\paragraph{Intervention.} There are several types of common \textit{interventions} which may be available \citep{eaton2007exact}. These are:
\textit{No intervention:} only observational data is obtained from the ground truth model. \textit{Hard/perfect:} the value of a single or several variables is fixed and then ancestral sampling is performed on the other variables. \textit{Soft/imperfect:} the conditional distribution of the variable on which the intervention is performed is changed. \textit{Uncertain:} the learner is not sure of which variable exactly the intervention affected directly. Here we make use of soft intervention because they include hard intervention as a limiting case and hence are more general.
\paragraph{Structure discovery using continuous optimization.} 
Structure discovery is a super-exponential search problem that searches though all possible directed acyclic graphs (DAGs). Previous continuous-optimization structure learning works \citep{zheng2018dags,yu2019dag,lachapelle2019gradient} mitigate the problem of searching in the super-exponential set of graph structures by considering the degree to which a hypothesis graph violates ``DAG-ness'' as an additional penalty to be optimized.
If there are $M$ such variables, the strategy of considering all the possible structural graphs as separate hypotheses is not feasible because it would require maintaining $O(2^{M^2})$ models of the data. 
\vspace{-2mm}
\section{Related Work}\label{related_work}
\vspace{-2mm}
The recovery of the underlying structural causal graph from observational and interventional data is a fundamental problem \citep{pearl1995causal, pearl2009causality,spirtes2000causation}. Different  approaches have been studied: score-based, constraint-based, asymmetry-based  and continuous optimization methods. Score-based methods search through the space of all possible directed acyclic graphs (DAGs) representing the causal structure based on some form of scoring function for network structures \citep{heckerman1995learning,chickering2002optimal,tsamardinos2006max,hauser2012characterization,goudet2017causal,cooper1999causal,zhu2019causal}. Constraint-based methods \citep{spirtes2000causation, sun2007kernel, zhang2012kernel,monti2019causal,zhu2019causal} infer the DAG by analyzing conditional independences in the data.
\citet{eatonuai} use dynamic programming techniques to accelerate Markov Chain Monte Carlo (MCMC) sampling in a Bayesian approach to structure learning for discrete variable DAGs. Asymmetry-based methods \citep{shimizu2006linear, hoyer2009nonlinear, Peters2011b, daniusis2012inferring, Budhathoki17, mitrovic2018causal} assume asymmetry between cause and effect in the data and try to use this information to estimate the causal structure.
\citet{peters2016causal, ghassami2017learning,rojas2018invariant}  exploit invariance across environments to infer causal structure, which faces difficulty scaling due to the iteration over the super-exponential set of possible graphs. \citet{mooij2016joint} propose a modelling framework that leverages existing methods while being more powerful and applicable to a wider range of settings. Recently, \citep{zheng2018dags,yu2019dag,lachapelle2019gradient} framed the structure search as a  continuous optimization problem, however, the methods only uses observational data and are non-trivial to extend to interventional data. In  our paper, we present a method  that uses continuous optimization methods that works on both observational and interventional data.

For interventional data, it is often assumed that the models have access to full intervention information, which is rare in the real world. \citet{rothenhausler2015backshift} have investigated the case of additive shift interventions, while \citet{eaton2007exact} have examined the situation where the targets of experimental interventions are imperfect or uncertain. This is different from our setting where the intervention is unknown to start with and is assumed to arise from other agents and the environment. \citet{bengio2019meta} propose a meta-learning framework for learning causal models from interventional data. However, the  method \cite{bengio2019meta} explicitly models every possible set of parents for every child variable and attempts to distinguish the best amongst the  combinatorially many such parent sets. It cannot scale beyond trivial graphs and only 2 variable experiments are presented in the paper.

 Learning based methods have been proposed \citep{guyoncause,guyonchalearn, lopez2015towards}
and there also exist recent approaches using the generalization ability of neural networks to learn causal signals from purely observational data \citep{kalainathan2018sam, goudet2018learning}. Neural network methods equipped with learned masks, such as \citep{ivanov2018variational, li2019flow, yoon2018gain, douglas2017universal}, exist in the literature, but only a few \citep{kalainathan2018sam} have been adapted to causal inference. This last work is, however, tailored for causal inference on continuous variables and from observations only. Adapting it to a discrete-variable setting is made difficult by its use of a Generative Adversarial Network (GAN) \cite{goodfellow2014generative} framework.
\vspace{-2mm}
\section{Structure Discovery from Interventions Method}\label{method}
\subsection{Scope of Applicability and Objective}
The proposed method, like any structure learning algorithm, assumes the availability of a data-generating process based on ancestral sampling of a black-box SCM of $M$ variables, which can be queried for samples. Because the method supports interventions, we further assume that the black-box supports applying and retracting interventions, although they may not be visible outside the black-box. The black-box can support infinite- or finite-data as well as infinite- or finite-intervention regimes.

The objective is, then, to "open up" the black-box, and learn the SCM's concealed edge structure from the glimpses that each intervention affords into the black-box's behaviour.

\subsection{Problem Setting and Assumptions}
In this paper, we restrict the problem setting to specific, but still broad classes of SCMs and interventions. In particular, we assume that:
\begin{itemize}
    \itemsep1pt
    \item \textit{Data is discrete-valued.} The SCM's random variables are all categorical.
    \item \textit{Data is fully observed.} For every sample, the value of all random variables are available.
    \item \textit{Interventions are sparse.} They affect only a single random variable (but which one may not be known). This is realistic because a large-scale, coordinated intervention on a broad subset of a realistic system's causal mechanisms is implausible for any single agent.
    \item \textit{Interventions are soft.} An intervention does not necessarily pin its target random variable to a fixed value (although it may).
    \item \textit{Interventions do not stack.} Before a new intervention is made, the previous one is fully retracted. This prevents the black-box's behaviour from wandering away from the observational configuration after a long series of interventions.
    \item \textit{No control over interventions.} The structure learning algorithm has control neither of the target, nor the nature of the next intervention inside the black box.
\end{itemize}
For a detailed description of the interventions, refer to \S\ref{appendix_intervention}. 
\vspace{-2mm}
\subsection{Variations and Prior Knowledge}\label{sec:variations}
In the problem setting above, the black-box is completely opaque. However, there exist two interesting relaxations of the black-box formulation to a more translucent, ``gray'' box:
\begin{itemize}
    \item \textbf{Complete or partial graph recovery:} We may already know the existence of certain cause-effect edges and non-edges within the black-box SCM. If such prior information is available, it may greatly speed up the search for the true causal graph, turning a \textit{complete graph recovery} problem into one of \textit{partial graph recovery}.
    \item \textbf{Known or unknown-target interventions:} The black-box might ``leak'' the target of an intervention. If so, we are not required to \textbf{predict} the target of the intervention.
\end{itemize}
We demonstrate that the proposed method can naturally incorporate this prior information to improve its performance.
\subsection{Method Overview}
\begin{wrapfigure}{r}{0.3\textwidth}
\vspace{-5\baselineskip}
  \begin{center}
    \includegraphics[width=0.3\textwidth]{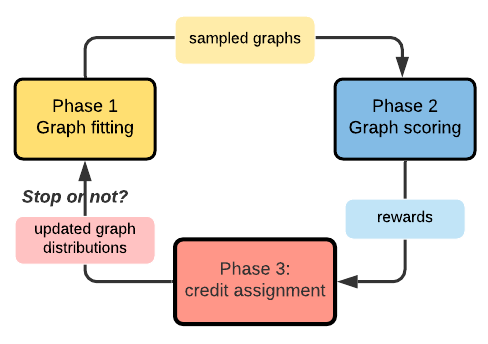}
  \end{center}
  \caption{ Workflow for our proposed method  SDI. Phase 1 samples graphs under the model's current belief about the edge structure and fits parameters to observational data. Phase 2 scores a small set of graphs against interventional data and assigns rewards according to graphs' ability to predict interventions. Phase 3 uses the rewards from Phase 2 to update the beliefs about the edge structure. If the believed edge probabilities have all saturated near 0 or 1, the method has converged. }
  \label{fig:framework}
  \vspace{-2.5\baselineskip}
\end{wrapfigure}
The proposed method is a \textit{score-based}, \textit{iterative}, \textit{continuous-optimization} method in three phases that flow into each other (See Figure \ref{fig:framework}). During the three-phase procedure, a structural representation of a DAG and a functional representation of a set of independent causal mechanisms are trained jointly until convergence. Because the structural and functional parameters are not independent and do influence each other, we train them in alternating phases, a form of block coordinate descent optimization.

\subsubsection{Parametrization}

We distinguish two sets of parameters: The \textit{structural parameters} $\gamma$ and the \textit{functional parameters} $\theta$. Given a graph of $M$ variables, we parametrize the structure $\gamma$ as a matrix $\mathbb{R}^{M \times M}$ such that $\sigmoid(\gamma_{ij})$ is our belief in random variable $X_j$ being a direct cause of $X_i$, where $\sigmoid(x) = 1/(1+\exp(-x))$ is the sigmoid function. The matrix $\sigmoid(\gamma)$ is thus a softened \textit{adjacency matrix}.

The set of all functional parameters $\theta$ comprises the parameters $\theta_i$ that model the conditional probability distribution of $X_i$ given its parent set $X_{\textrm{pa}(i, C)}$, with $C \sim \mathrm{Ber}(\sigma(\gamma))$ a hypothesized configuration of the SCM's DAG.

\subsubsection{Phase 1: Graph Fitting on Observational Data}\label{phase1}

During Phase 1, the functional parameters $\theta$ are trained to maximize the likelihood of randomly drawn observational data under graphs randomly drawn from our current beliefs about the edge structure. We draw graph configurations $C_{ij} \sim \mathrm{Ber}(\sigma(\gamma_{ij}))$ and batches of observational data from the unintervened black-box SCM, then maximize the log-likelihood of the batch under that configuration using SGD. The use of graph configurations sampling from Bernoulli distributions is analogous to dropout on the inputs of the functional models (in our implementation, MLPs), giving us an ensemble of neural networks that can model the observational data. 
\subsubsection{Phase 2: Graph Scoring on Interventional Data}\label{phase2}
During Phase 2, a number of graph configurations are sampled from the current edge beliefs parametrized by $\gamma$, and scored on data samples drawn from the intervened black-box SCM.

\textbf{Intervention applied:} At the beginning of Phase 2, an intervention is applied to the black-box SCM. This intervention is not under the control of the method. In our implementation, and unbeknownst to the model, the target variable is chosen uniformly randomly from all $M$ variables throughout the optimization process.

\textbf{Intervention predicted:} If the target of the intervention is not known, it is \textit{predicted} using a simple heuristic. A small number of interventional data samples are drawn from the black-box and more graphs are sampled from our current edge beliefs. The average log-likelihood of each individual variable $X_i$ across the samples is then computed using the functional model parameters $\theta$ fine-tuned on observational data in Phase 1. The variable $X_i$ showing the greatest deterioration in log-likelihood is assumed to be the target because the observational distribution most poorly predicts that variable.

If the target of the intervention leaks from the black-box, then this is taken as ground-truth knowledge for the purpose of subsequent steps, and no prediction is done.

\textbf{Graphs Sampled and Scored:} A new set of interventional data samples and graph configurations are now drawn from the intervened black-box and edge beliefs respectively. The log-likelihood of the data batches under the hypothesized configurations is computed, with one modification: The contribution to the total log-likelihood of a sample $X$ coming from the intervened (or predicted-intervened) random variable $X_i$ is masked.
Because $X_i$ was intervened upon (in the manner of a Pearl do-operation, soft or hard), the values one gets for that variable should be taken as givens, not as contributors to the total log-likelihood of the sample, and nor should a correction gradient propagate through it (because the variable's CPT or MLP wasn't actually responsible for the outcome).

\textbf{Intervention retracted:} After Phase 2, the intervention is retracted, per our modelling assumptions.
\subsubsection{Phase 3: Credit Assignment to Structural Parameters}\label{phase3}
During Phase 3, the scores of the interventional data batches over various graph configurations are aggregated into a gradient for the structural parameters $\gamma$. Because a discrete Bernoulli random sampling process was used to sample graph configurations under which the log-likelihoods were computed, we require a gradient estimator to propagate gradient through to the $\gamma$ structural parameters. Several alternatives exist, but we adopt for this purpose the REINFORCE-like gradient estimator $g_{ij}$ with respect
to $\gamma_{ij}$ proposed by \citet{bengio2019meta}:
\vspace{-2mm}
\begin{align}
    g_{ij} &= \frac{\sum_{k} ( \sigma(\gamma_{ij}) - c^{(k)}_{ij} ) \mathcal{L}_{C\makebox[0mm][l]{\tiny\raisebox{1.2mm}[0mm][0mm]{\hspace{-0.5mm}$^{(k)}$}},i}\,(X) }{\sum_k \mathcal{L}_{C\makebox[0mm][l]{\tiny\raisebox{1.2mm}[0mm][0mm]{\hspace{-0.5mm}$^{(k)}$}},i}\,(X)}, \quad \forall i,j \in \{0, \ldots, M\!-\!1\} \label{eq:metatransferloss}
\end{align}
where the $\smash{^{(k)}}$ superscript indicates the values obtained for the $k$-th draw of $C$ under the current edge beliefs parametrized by $\gamma$. Therefore, $\smash{\mathcal{L}_{C,i}^{(k)}(X)}$ can be read as the log-likelihood of variable $X_i$ in the data sample $X$ under the $k$'th configuration, $\smash{C^{(k)}}$, drawn from our edge beliefs. Using the estimated gradient, we then update $\gamma$ with SGD, and return to Phase 1 of the continuous optimization process.

\textbf{Acyclic Constraint:} We include a regularization term $J_\textrm{DAG}(\gamma)$ that penalizes length-2 cycles in the learned adjacency matrix $\sigmoid(\gamma)$, with a tunable strength $\lambda_\textrm{DAG}$. The regularization term is
$J_\textrm{DAG}(\gamma) = \sum_{i \ne j} \cosh(\sigma(\gamma_{ij})\sigma(\gamma_{ji})), \quad \forall i,j \in \{0, \ldots, M\!-\!1\}$
and is derived from \citet{zheng2018dags}. The details of the derivation are in the Appendix. We explore several different values of $\lambda_\textrm{DAG}$ and their effects in our experimental setup. Suppression of longer-length cycles was not found to be worthwhile for the increased computational expense.
\vspace{-2mm}
\section{Experimental Setup and Results}
We first evaluate the proposed method on a synthetic dataset where we have control over the number of variables and causal edges in the ground-truth SCM. This allows us to analyze the performance of the proposed method under various conditions. We then evaluate the proposed method on real-world datasets from the BnLearn dataset repository. We also consider the two variations of \S\ref{sec:variations}: Recovering only part of the graph (when the rest is known), and exploiting knowledge of the intervention target.

The summary of our findings is: 1) We show strong results for graph recovery for all synthetic graphs in comparisons with other baselines, measured by Hamming distance. 2) The proposed method achieves high accuracy on partial graph recovery for large, real-world graphs. 3) The proposed method's intervention target prediction heuristic closes the gap between the known- and unknown-target intervention scenarios. 4) The proposed method generalizes well to unseen interventions. 5) The proposed method's time-to-solution scaling appears to be driven by the number of edges in the groundtruth graph moreso than the number of variables.
\begin{wrapfigure}{r}{0.5\textwidth}
\centering
\includegraphics[width=0.5\textwidth]{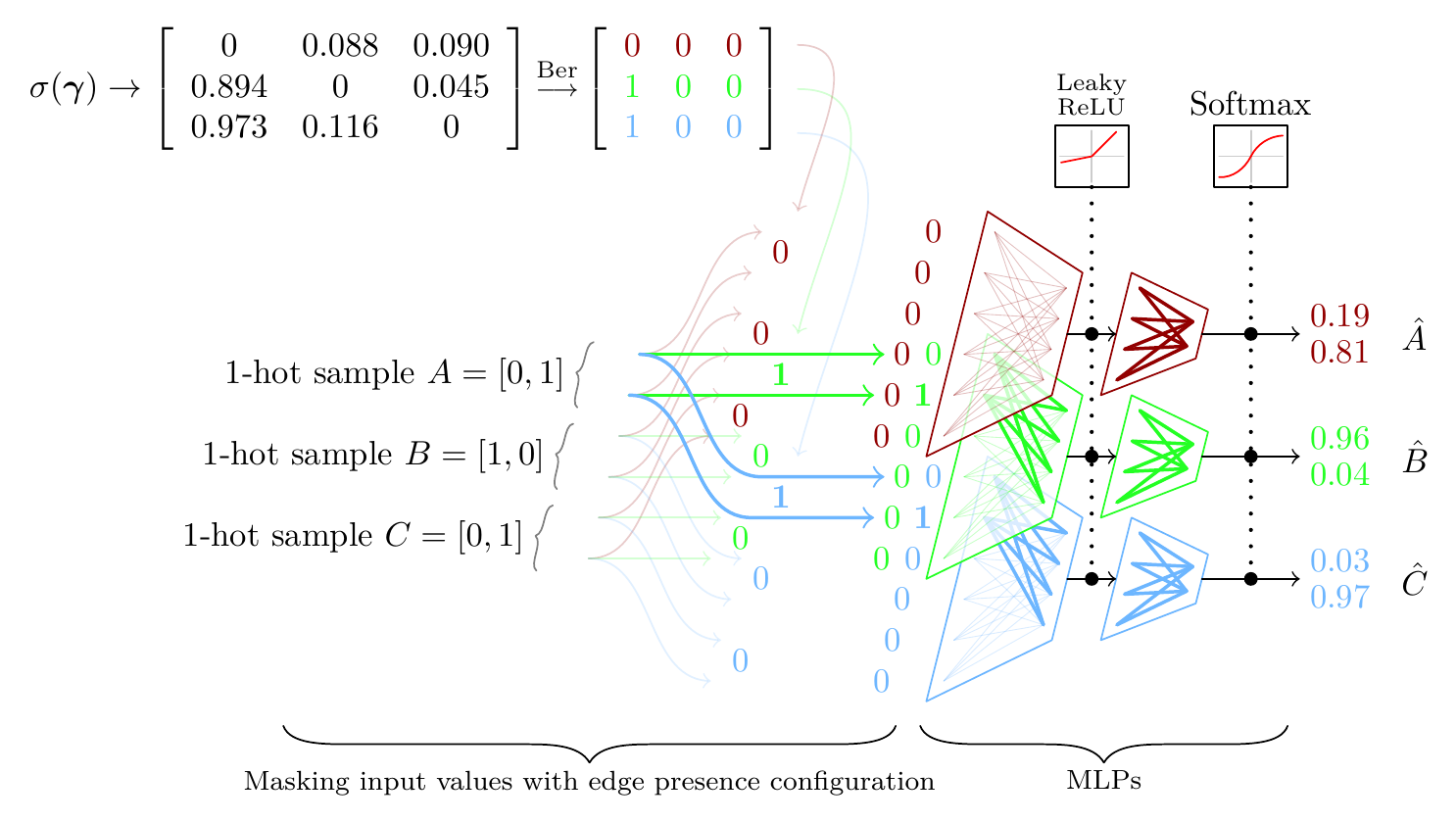}
\caption{MLP Model Architecture for $M=3$, $N=2$ (\texttt{fork3}) SCM. The model computes the conditional probabilities of 
$A, B, C$
given their parents using a stack of three independent MLPs. The MLP input layer uses an adjacency matrix sampled from $\mathrm{Ber(\sigma(\gamma))}$ as an input mask to force the model to make use only of parent nodes to predict their child node.}
\label{model_figure}
\vspace{-1.5\baselineskip}
\end{wrapfigure}
\vspace{-3mm}
\subsection{Model Description} \label{sec:modeldescription}
\vspace{-1mm}
\paragraph{Learner model.} Without loss of generality, we let $\theta_i = \{\texttt{W0}_i, \texttt{B0}_i, \texttt{W1}_i, \texttt{B1}_i\}$ define a stack of $M$ one-hidden-layer MLPs, one for each random variable $X_i$. A more appropriate model, such as a CNN, can be chosen using domain-specific knowledge; the primary advantage of using MLPs is that the hypothesized DAG configurations $c_{ij}$ can be readily used to mask the inputs of MLP $i$, as shown in Figure \ref{model_figure}.

To force the structural equation $f_i$ corresponding to $X_i$ to rely exclusively on its direct ancestor set $\textrm{pa}(i,C)$ under hypothesis adjacency matrix $C$ (See Eqn.~\ref{eq:structuralassignment}), the one-hot input vector $X_j$ for variable $X_i$'s MLP is masked by the Boolean element $c_{ij}$. An example of the multi-MLP architecture with $M$=3 categorical variables of $N$=2 categories is shown in Figure \ref{model_figure}. For more details, refer to Appendix \ref{appendix_modelsetup}.
\vspace{-1mm}
\paragraph{Ground-truth model.}
\vspace{-2mm}
Ground-truth SCM models are parametrized either as CPTs with parameters from BnLearn (in the case of real-world graphs), or as a second stack of MLPs similar to the learner model, with randomly-initialized functional parameters $\theta_\textrm{GT}$ and the desired adjacency matrix $\gamma_\textrm{GT}$.
\vspace{-3mm}
\paragraph{Interventions.} In all experiments, at most one (soft) intervention is concurrently performed. To simulate a soft intervention on variable $X_i$, we reinitialize its ground-truth conditional distribution's MLP parameters or CPT table randomly, while leaving the other variables untouched. For more details about the interventions, please refer to Appendix \ref{appendix_intervention}.
\vspace{-2mm}
\subsection{Synthetic Datasets Experiments}\label{synthetic_data}
\vspace{-3mm}
We first evaluate the model's performance on several randomly-initialized SCMs with specific, representative graph structures. Since the number of possible DAGs grows super-exponentially with the number of variables, for $M$=4 up to 13 a selection of representative and edge-case DAGs are chosen. \texttt{chainM} and \texttt{fullM} ($M$=3-13) are the minimally- and maximally-connected $M$-variable DAGs, while \texttt{treeM} and \texttt{jungleM} are tree-like intermediate graphs. \texttt{colliderM} is the $(M\!-\!1) \to 1$ collider graph. The details of the setup is in Appendix \ref{exp_setup_synthetic}.
\begin{wrapfigure}{r}{0.8\textwidth}
    \centering
    \includegraphics[scale=0.35]{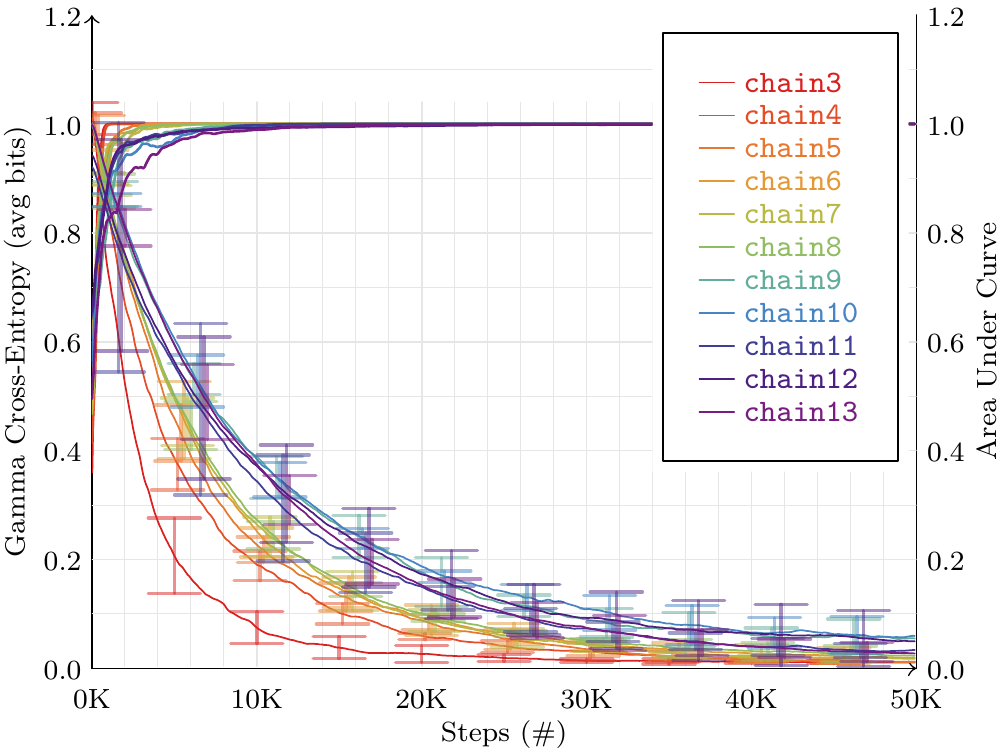}
    \includegraphics[scale=0.35]{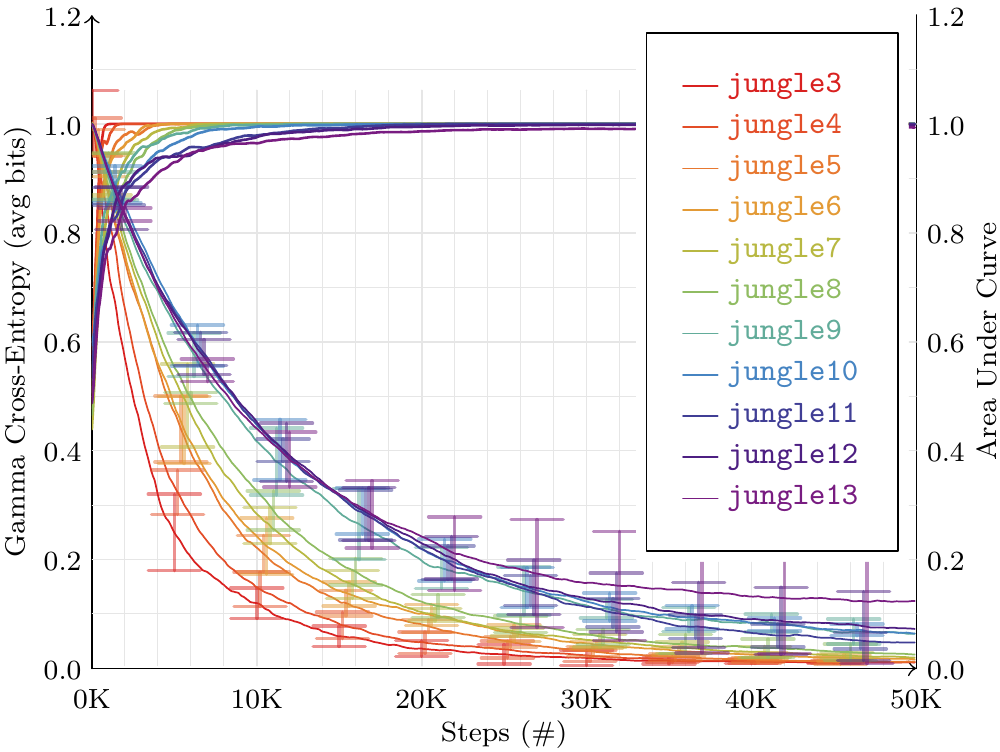}
    \includegraphics[scale=0.35]{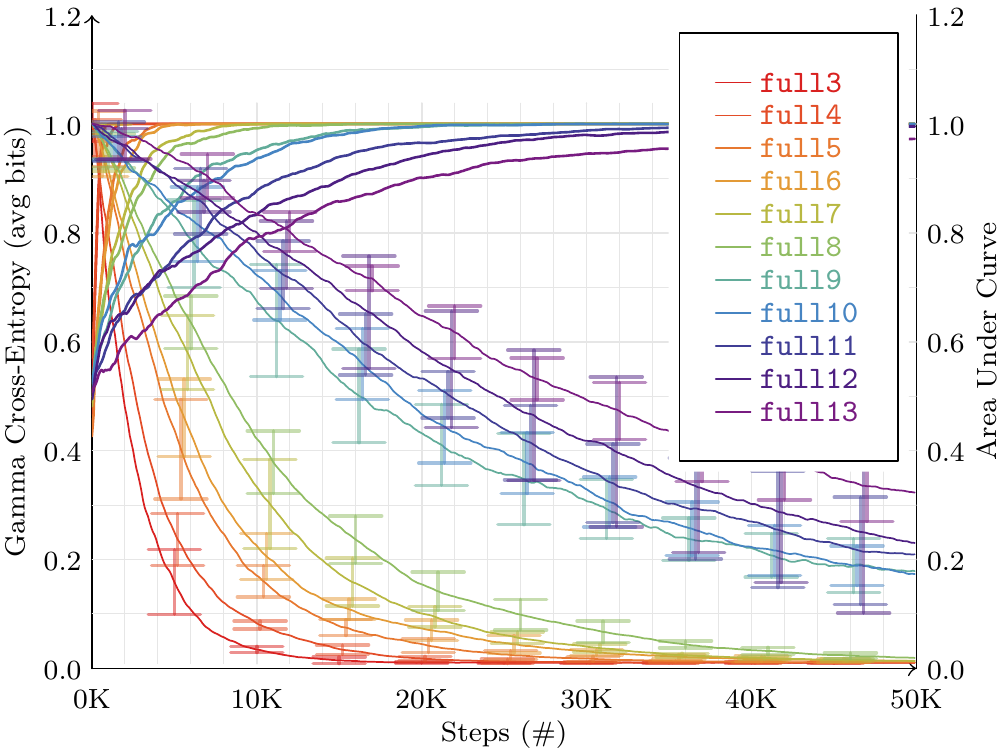}
    \caption{Cross entropy (CE) and Area-Under-Curve (AUC/AUROC) for edge probabilities of learned graph against ground-truth for synthetic SCMs. Error bars represent $\pm 1\sigma$ over PRNG seeds 1-5. \textbf{Left to right}: 
    \texttt{chainM},\texttt{jungleM},\texttt{fullM},$M=3\ldots13$. Graphs (3-13 variables) all learn perfectly with AUROC reaching 1.0. However, denser graphs (fullM) take longer to converge.}
    \label{fig:synthetic_data_results}
        \vspace{-2\baselineskip}
\end{wrapfigure}
\paragraph{Results.} The model can recover most synthetic DAGs with high accuracy, as measured by Structural Hamming Distance (SHD) between learned and ground-truth DAGs. Table \ref{table:all_baseline_hamming} shows our proposed method outperforming all other baseline methods, and learns all graphs perfectly for 3 to 13 variables (excepting \texttt{full}). For DAGs ranging from 3 to 8 variables, the AUROCs all eventually reach 1.0 (indicating perfect classification into edge/non-edge; Refer to Figure \ref{fig:synthetic_data_results}). For both large ($M>10$) and dense DAGs (e.g.~\texttt{full13}) the model begins encountering difficulties, as shown in Table \ref{table:all_baseline_hamming} and Appendix \S\ref{appendix_synthetic_results}.

Small graphs ($M<10$) are less sensitive than larger ones to our hyperparameters, notably the sparsity and acyclic regularization (\S\ref{phase3}) terms. In \S\ref{appendix_hyperparam}, we perform an analysis of these hyperparameters.
\vspace{-2mm}
\begin{table*}[h]
 \centering  
\caption{\textbf{Baseline  comparisons:} Structural Hamming Distance (SHD) (lower is better) for learned and ground-truth edges on various graphs from both synthetic and real datasets, compared to \citep{peters2016causal}, \citep{heinze2018invariant}, \citep{eaton2007exact}, \citep{yu2019dag} and \citep{zheng2018dags}. The proposed method (\gls{SDI}) is run on random seeds $1-5$ and we pick the worst performing model out of the random seeds in the table. OOM: out of memory. Our proposed method correctly recovers the true causal graph, with the exception of Sachs and full13, and it significantly outperforms all other baseline methods.}
\begin{tabular}{lrrrrrrr}
 \toprule
{\bf \small Method}  & \texttt{\small Asia} & \texttt{\small  Sachs} &  \texttt{ \small  collider}   & \texttt{\small chain}   & \texttt{\small jungle} & \texttt{\small collider}& \texttt{\small full} \\
$M$ & \texttt{\small 8} & \texttt{\small 11} & \texttt{\small 8} & \texttt{\small 13} & \texttt{\small 13} & \texttt{\small 13} & \texttt{\small 13} \\
 \midrule
{\bf \small ~\citet{zheng2018dags}}       &  14  & 22  & 18 & 39    &   22   &   24 &   71   \\
{\bf \small ~\citet{yu2019dag}}           &  10  &  19  & 7 &   14   &  16     &     12  & 77  \\
{\bf \small ~\citet{heinze2018invariant} } &   8  &  17 &  7   &  12   &   12       &  7 &    28           \\
{\bf \small ~\citet{peters2016causal}}    &   5  & 17  & 2 &  2   &        8            &    2  &   16            \\
{\bf \small ~\citet{eaton2007belief}}      &   0  &  OOM &   7   &    OOM     &   OOM           &    OOM  &        OOM      \\
{\bf \small Proposed Method (\gls{SDI})}                  &   \highlight{0}  &   \highlight{6}  & \highlight{0}  & \highlight{0}  &   \highlight{0}    &    \highlight{0} &    \highlight{7}  \\
 \bottomrule
 \end{tabular}
 \vspace{-1\baselineskip}
 \label{table:all_baseline_hamming}
\end{table*}
\subsection{Real-World Datasets: BnLearn}\label{bnlearn} 
The Bayesian Network Repository is a collection of commonly-used causal Bayesian networks from the literature, suitable for Bayesian and causal learning benchmarks.  We evaluate the proposed method on the Earthquake \citep{korb2010bayesian}, Cancer \citep{korb2010bayesian},  Asia \citep{lauritzen1988local} and Sachs \citep{sachs2005causal} datasets ($M=$5, 5, 8 and 11-variables respectively, maximum in-degree 3) in the BnLearn dataset repository.

\paragraph{Results.} As shown in Table \ref{table:all_baseline_hamming}, the proposed method perfectly recovers the DAG of Asia, while making a small number of errors (SHD=6) for Sachs (11-variables). It thus significantly outperforms all other baselines models. Figures  \ref{fig:gamma:synthetic_results} \& \ref{fig:gamma:earthquake} visualize what the model has learned at several stages of learning. Results for Cancer and Asia can be found in the appendices, Figure \ref{fig:cancer_temp} and \ref{fig:earthquake_CE}.
\vspace{-2mm}
 \subsection{Comparisons with other methods}\label{baseline_compare}
 \vspace*{-1mm}
As shown in Table \ref{table:all_baseline_hamming}, we compared the proposed \gls{SDI} method to  ICP~(\citep{peters2016causal}), non-linear ICP~(\citep{heinze2018invariant}), and \citep{eaton2007exact,zheng2018dags,yu2019dag} on Asia~\citep{lauritzen1988local}, Sachs ~\citep{sachs2005causal} and representative synthetic graphs. \citet{eaton2007exact} handles uncertain interventions and \citet{peters2016causal}, \citet{heinze2018invariant} handles unknown interventions. However, neither attempts to predict the intervention. As shown in Table  \ref{table:all_baseline_hamming}, we significantly outperform ICP, non-linear ICP, and the methods in \citep{yu2019dag} and \citep{zheng2018dags}. Furthermore, \citet{eaton2007exact} runs out of memory for graphs larger than $M=10$ because modelling of uncertain interventions is done using ``shadow'' random variables (as suggested by the authors), and thus recovering the DAG internally requires solving a $d=2M$-variable problem. Their method's extremely poor time- and space-scaling of $O(d 2^d)$ makes it unusable beyond $d > 20$.

For \gls{SDI}s, we threshold our edge beliefs at $\sigmoid(\gamma) = 0.5$ to derive a graph, but the continued decrease of the cross-entropy loss (Figure \ref{fig:synthetic_data_results}) hints at \gls{SDI}'s convergence onto the correct causal model. Please refer to Appendix \S\ref{annex:comparison} for full details and results.
\vspace{-2mm}

\vspace*{-1mm}
\subsection{Generalization to Previously Unseen Interventions}\label{generalization}
\vspace*{-1mm}
\begin{wraptable}{r}{8.5cm}
\centering  
\vspace{-1\baselineskip}
\caption{\textbf{Evaluating the consequences of a previously unseen intervention:} (test log-likelihood under intervention)}
\scalebox{0.90}{{
\begin{tabular}{ccccc}
  \toprule
    & fork3   & { chain3}  & { confounder3}&{{collider3}}\\
  \midrule
    {\bf Baseline}   & -0.5036  &  -0.4562 & -0.3628 & -0.5082 \\
    {\bf \gls{SDI}}  &  \highlight{-0.4502}  & \highlight{-0.3801} & \highlight{-0.2819} & \highlight{-0.4677}\\
  \bottomrule
\end{tabular}
}}
\label{tab:generalization}
\vspace{-1\baselineskip}
\end{wraptable}
It is often argued that machine learning approaches based purely on capturing joint distributions do not necessarily yield models that generalize to unseen experiments, since they do not explicitly model changes through interventions. By way of contrast, causal models use the concept of interventions to explicitly model changing environments and thus hold the promise of robustness even under distributional shifts \citep{pearl2009causality, SchJanPetSgoetal12, peters2017elements}.  To test the robustness of causal modelling to previously unseen interventions (new values for an intervened variable), we evaluate a well-trained causal model against a variant, non-causal model trained with $c_{ij} = 1,\; i \ne j$. An intervention is performed on the black-box SCM, fresh interventional data is drawn from it, and the models, with knowledge of the intervention target, are asked to predict the other variables given their parents. The average log-likelihoods of the data under both models are computed and contrasted. The intervention variable's contribution to the log-likelihood is masked.
For all 3-variable graphs (\texttt{chain3}, \texttt{fork3}, \texttt{collider3}, \texttt{confounder3}), the causal model attributes higher log-likelihood to the intervention distribution's samples than the non-causal variant, thereby demonstrating causal models' superior generalization ability in transfer tasks. Table \ref{tab:generalization} collects these results.
\vspace{-1mm}

\vspace*{-1mm}
\subsection{Variant: Predicting interventions}\label{prediction}
\vspace*{-1mm}
In Phase 2 (\S\ref{phase2}), we use a simple heuristic to predict the intervention target variable. Experiments show that this heuristic functions well in practice, yielding correct predictions far more often than by chance alone (Table \ref{tab:predacc}). Guessing the intervention variable randomly, or not guessing it at all, leads to a significant drop in the model performance, even for 3-variable graphs (Figure \ref{fig:predict-or-not} Left). Training \gls{SDI} with intervention prediction closely tracks training with leaked knowledge of the ground-truth intervention on larger, 7-variable graphs (Figure \ref{fig:predict-or-not} Right).

\begin{wraptable}{r}{7.5cm}
\vspace{-2\baselineskip}
\centering  
\caption{\textbf{Intervention Prediction Accuracy:} (identify on which variable the intervention took place)}
{\begin{tabular}{llll}
\toprule
{\bf \small 3 variables}  & {\bf \small 4 variables}  & {\bf \small 5 variables} & {\bf \small 8 variables} \\
\midrule
95 \% & 93 \%  & 85 \%  & 71 \% \\
\bottomrule
\end{tabular} \\
}
\label{tab:predacc}
\end{wraptable}

\vspace*{-1mm}
\subsection{Variant: Partial Graph Recovery}\label{partial)_graph}
\vspace*{-1mm}
Instead of learning causal structures \textit{de novo}, we may have partial information about the black-box SCM and may only need to fill in missing information (\S\ref{sec:variations}). An example is protein structure discovery in biology, where some causal relationships have been definitely established and others remain open hypotheses. This is an easier task compared to full graph recovery, since the model only has to search for missing edges.
\begin{wraptable}{r}{5.5cm}
\vspace{-1\baselineskip}
\centering
\caption{\textbf{Partial Graph Recovery} on Alarm \citep{beinlich1989alarm} and Barley \citep{kristensen2002use}. The model is asked to predict 50 edges in Barley and 40 edges in Alarm. The accuracy is measured in Structural Hamming Distance (SHD). \gls{SDI} achieved over $90\%$ accuracy on both graphs.}
{\begin{tabular}{lrr}
\toprule
Graph & \textbf{Alarm}  & \textbf{Barley} \\
{\small Number of variables}             & 37 & 48 \\
{\small Total Edges} & 46 & 84 \\
\midrule
Edges to recover & 40 & 50 \\
Recovered Edges & 37 & 45 \\
Errors (in SHD)         &  \highlight{3} &  \highlight{5} \\
\bottomrule
\end{tabular} \\
}
\vspace{-2\baselineskip}
\label{tab:partial_graph}
\end{wraptable}
We evaluate the proposed method on Barley \citep{kristensen2002use} ($M=48$) and Alarm \citep{beinlich1989alarm} ($M=37$) from the BnLearn repository. The model is asked to predict 50 edges from Barley and 40 edges from Alarm. The model reached $\ge 90\%$ accuracy on both datasets, as shown in Table \ref{tab:partial_graph}.

\vspace*{-1mm}
\subsection{Ablation and analysis}
\vspace{-2mm}
As shown in Figure \ref{fig:sparsity}, larger graphs (such as $M>6$) and denser graphs (such as \texttt{full8}) are progressively more difficult to learn. For denser graphs, the learned models have higher sample complexity, higher variance and slightly worse results. Refer to Appendix \S\ref{appendix_sparsity} for complete results on all graphs.
\vspace{-2mm}
\paragraph{Hyperparameters.}
Hyperparameters for all experiments were kept identical unless otherwise stated. We study the effect of DAG and sparsity penalties in the following paragraph. For more details on hyperparameter setup, please refer to Appendix \S\ref{appendix_hyperparam} . 
\vspace{-1mm}
\paragraph{Importance of regularization.} Valid configurations $C$ for a causal model are expected to be a) sparse and b) acyclic. To promote such solutions, we introduce DAG and sparsity regularization with tunable hyperparameters. For all experiments, we set the DAG penalty to $0.5$ and sparsity penalty to $0.1$. We run ablation studies on different values of the regularizers and study their effect. We find that smaller graphs are less sensitive to different values of regularizer than larger graphs. For details, refer to Appendix \S\ref{appendix_regularization}.
\vspace{-2mm}
\paragraph{Importance of dropout.}
To train functional parameter for an observational distribution, sampling adjacency matrices is required. We "drop out" each edge (with a probability of $\sigma(\gamma)$) in our experiments during functional parameter training of the conditional distributions of the SCM. Please refer to Appendix \S\ref{appendix_dropout} for a more detailed analysis.
\vspace{-2mm}
\section{Conclusion}
\vspace*{-1mm}
In this work, we introduced an experimentally successful method (\gls{SDI}) for causal structure discovery using continuous optimization, combining information from both observational and interventional data. We show in experiments that 
it can recover true causal structure, that it generalizes well to unseen interventions, that
it compares very well against the start-of-the-art causal discovery methods 
on real world datasets, and that it scales even better on problems
where only part of the graph is known.

\section*{Acknowledgements}
The authors would like to acknowledge the support of the following agencies for research funding and computing support: NSERC, Compute Canada, the Canada Research Chairs, CIFAR, and Samsung. We would also like to thank the developers of Pytorch for developments of great frameworks. We wish to thank Sébastien Lachapelle for referring us to \cite{zheng2018dags} and help to extend the acyclic regularization term $J_{DAG}$ to the current from. We would like to thank Lars Buesing, Bernhard Schölkopf, Nasim Rahaman, Jovana Mitrović and Rémi Le Priol for useful feedback and discussions.

\bibliographystyle{unsrtnat}
\bibliography{main}

\begin{thebibliography}{56}
\providecommand{\natexlab}[1]{#1}
\providecommand{\url}[1]{\texttt{#1}}
\expandafter\ifx\csname urlstyle\endcsname\relax
  \providecommand{\doi}[1]{doi: #1}\else
  \providecommand{\doi}{doi: \begingroup \urlstyle{rm}\Url}\fi

\bibitem[Pearl and Mackenzie(2018)]{pearl2018book}
Judea Pearl and Dana Mackenzie.
\newblock \emph{The book of why: the new science of cause and effect}.
\newblock Basic Books, 2018.

\bibitem[Sachs et~al.(2005)Sachs, Perez, Pe'er, Lauffenburger, and
  Nolan]{sachs2005causal}
Karen Sachs, Omar Perez, Dana Pe'er, Douglas~A Lauffenburger, and Garry~P
  Nolan.
\newblock Causal protein-signaling networks derived from multiparameter
  single-cell data.
\newblock \emph{Science}, 308\penalty0 (5721):\penalty0 523--529, 2005.

\bibitem[Hill et~al.(2016)Hill, Heiser, Cokelaer, Unger, Nesser, Carlin, Zhang,
  Sokolov, Paull, Wong, et~al.]{hill2016inferring}
Steven~M Hill, Laura~M Heiser, Thomas Cokelaer, Michael Unger, Nicole~K Nesser,
  Daniel~E Carlin, Yang Zhang, Artem Sokolov, Evan~O Paull, Chris~K Wong,
  et~al.
\newblock Inferring causal molecular networks: empirical assessment through a
  community-based effort.
\newblock \emph{Nature methods}, 13\penalty0 (4):\penalty0 310--318, 2016.

\bibitem[Abramson et~al.(1996)Abramson, Brown, Edwards, Murphy, and
  Winkler]{abramson1996hailfinder}
Bruce Abramson, John Brown, Ward Edwards, Allan Murphy, and Robert~L Winkler.
\newblock Hailfinder: A bayesian system for forecasting severe weather.
\newblock \emph{International Journal of Forecasting}, 12\penalty0
  (1):\penalty0 57--71, 1996.

\bibitem[Lauritzen and Spiegelhalter(1988)]{lauritzen1988local}
Steffen~L Lauritzen and David~J Spiegelhalter.
\newblock Local computations with probabilities on graphical structures and
  their application to expert systems.
\newblock \emph{Journal of the Royal Statistical Society: Series B
  (Methodological)}, 50\penalty0 (2):\penalty0 157--194, 1988.

\bibitem[Korb and Nicholson(2010)]{korb2010bayesian}
Kevin~B Korb and Ann~E Nicholson.
\newblock \emph{Bayesian artificial intelligence}.
\newblock CRC press, 2010.

\bibitem[Spirtes et~al.(2000)Spirtes, Glymour, Scheines, Heckerman, Meek,
  Cooper, and Richardson]{spirtes2000causation}
Peter Spirtes, Clark~N Glymour, Richard Scheines, David Heckerman, Christopher
  Meek, Gregory Cooper, and Thomas Richardson.
\newblock \emph{Causation, prediction, and search}.
\newblock MIT press, 2000.

\bibitem[Shimizu et~al.(2006)Shimizu, Hoyer, Hyv{\"a}rinen, and
  Kerminen]{shimizu2006linear}
Shohei Shimizu, Patrik~O Hoyer, Aapo Hyv{\"a}rinen, and Antti Kerminen.
\newblock A linear non-gaussian acyclic model for causal discovery.
\newblock \emph{Journal of Machine Learning Research}, 7\penalty0
  (Oct):\penalty0 2003--2030, 2006.

\bibitem[Heckerman et~al.(1995)Heckerman, Geiger, and
  Chickering]{heckerman1995learning}
David Heckerman, Dan Geiger, and David~M Chickering.
\newblock Learning bayesian networks: The combination of knowledge and
  statistical data.
\newblock \emph{Machine learning}, 20\penalty0 (3):\penalty0 197--243, 1995.

\bibitem[Eaton and Murphy(2007{\natexlab{a}})]{eaton2007belief}
Daniel Eaton and Kevin Murphy.
\newblock Belief net structure learning from uncertain interventions.
\newblock \emph{J Mach Learn Res}, 1:\penalty0 1--48, 2007{\natexlab{a}}.

\bibitem[Mooij et~al.(2016)Mooij, Magliacane, and Claassen]{mooij2016joint}
Joris~M Mooij, Sara Magliacane, and Tom Claassen.
\newblock Joint causal inference from multiple contexts.
\newblock \emph{arXiv preprint arXiv:1611.10351}, 2016.

\bibitem[Tillman and Spirtes(2011)]{tillman2011learning}
Robert Tillman and Peter Spirtes.
\newblock Learning equivalence classes of acyclic models with latent and
  selection variables from multiple datasets with overlapping variables.
\newblock In \emph{Proceedings of the Fourteenth International Conference on
  Artificial Intelligence and Statistics}, pages 3--15, 2011.

\bibitem[Rothenh{\"a}usler et~al.(2015)Rothenh{\"a}usler, Heinze, Peters, and
  Meinshausen]{rothenhausler2015backshift}
Dominik Rothenh{\"a}usler, Christina Heinze, Jonas Peters, and Nicolai
  Meinshausen.
\newblock Backshift: Learning causal cyclic graphs from unknown shift
  interventions.
\newblock In \emph{Advances in Neural Information Processing Systems}, pages
  1513--1521, 2015.

\bibitem[Zheng et~al.(2018)Zheng, Aragam, Ravikumar, and Xing]{zheng2018dags}
Xun Zheng, Bryon Aragam, Pradeep~K Ravikumar, and Eric~P Xing.
\newblock {DAGs} with {NO TEARS}: Continuous optimization for structure
  learning.
\newblock In \emph{Advances in Neural Information Processing Systems}, pages
  9472--9483, 2018.

\bibitem[Yu et~al.(2019)Yu, Chen, Gao, and Yu]{yu2019dag}
Yue Yu, Jie Chen, Tian Gao, and Mo~Yu.
\newblock Dag-gnn: Dag structure learning with graph neural networks.
\newblock \emph{arXiv preprint arXiv:1904.10098}, 2019.

\bibitem[Peters et~al.(2017)Peters, Janzing, and
  Sch{\"o}lkopf]{peters2017elements}
Jonas Peters, Dominik Janzing, and Bernhard Sch{\"o}lkopf.
\newblock \emph{Elements of causal inference: foundations and learning
  algorithms}.
\newblock MIT press, 2017.

\bibitem[Eberhardt et~al.(2012)Eberhardt, Glymour, and
  Scheines]{eberhardt2012number}
Frederick Eberhardt, Clark Glymour, and Richard Scheines.
\newblock On the number of experiments sufficient and in the worst case
  necessary to identify all causal relations among n variables.
\newblock \emph{arXiv preprint arXiv:1207.1389}, 2012.

\bibitem[Eaton and Murphy(2007{\natexlab{b}})]{eaton2007exact}
Daniel Eaton and Kevin Murphy.
\newblock Exact bayesian structure learning from uncertain interventions.
\newblock In \emph{Artificial Intelligence and Statistics}, pages 107--114,
  2007{\natexlab{b}}.

\bibitem[Lachapelle et~al.(2019)Lachapelle, Brouillard, Deleu, and
  Lacoste-Julien]{lachapelle2019gradient}
S{\'e}bastien Lachapelle, Philippe Brouillard, Tristan Deleu, and Simon
  Lacoste-Julien.
\newblock Gradient-based neural dag learning.
\newblock \emph{arXiv preprint arXiv:1906.02226}, 2019.

\bibitem[Pearl(1995)]{pearl1995causal}
Judea Pearl.
\newblock Causal diagrams for empirical research.
\newblock \emph{Biometrika}, 82\penalty0 (4):\penalty0 669--688, 1995.

\bibitem[Pearl(2009)]{pearl2009causality}
Judea Pearl.
\newblock \emph{Causality}.
\newblock Cambridge university press, 2009.

\bibitem[Chickering(2002)]{chickering2002optimal}
David~Maxwell Chickering.
\newblock Optimal structure identification with greedy search.
\newblock \emph{Journal of machine learning research}, 3\penalty0
  (Nov):\penalty0 507--554, 2002.

\bibitem[Tsamardinos et~al.(2006)Tsamardinos, Brown, and
  Aliferis]{tsamardinos2006max}
Ioannis Tsamardinos, Laura~E Brown, and Constantin~F Aliferis.
\newblock The max-min hill-climbing bayesian network structure learning
  algorithm.
\newblock \emph{Machine learning}, 65\penalty0 (1):\penalty0 31--78, 2006.

\bibitem[Hauser and B{\"u}hlmann(2012)]{hauser2012characterization}
Alain Hauser and Peter B{\"u}hlmann.
\newblock Characterization and greedy learning of interventional markov
  equivalence classes of directed acyclic graphs.
\newblock \emph{Journal of Machine Learning Research}, 13\penalty0
  (Aug):\penalty0 2409--2464, 2012.

\bibitem[Goudet et~al.(2017)Goudet, Kalainathan, Caillou, Guyon, Lopez-Paz, and
  Sebag]{goudet2017causal}
Olivier Goudet, Diviyan Kalainathan, Philippe Caillou, Isabelle Guyon, David
  Lopez-Paz, and Mich{\`e}le Sebag.
\newblock Causal generative neural networks.
\newblock \emph{arXiv preprint arXiv:1711.08936}, 2017.

\bibitem[Cooper and Yoo(1999)]{cooper1999causal}
Gregory~F. Cooper and Changwon Yoo.
\newblock Causal {Discovery} from a {Mixture} of {Experimental} and
  {Observational} {Data}.
\newblock In \emph{Proceedings of the {Fifteenth} {Conference} on {Uncertainty}
  in {Artificial} {Intelligence}}, {UAI}'99, pages 116--125, San Francisco, CA,
  USA, 1999.

\bibitem[Zhu and Chen(2019)]{zhu2019causal}
Shengyu Zhu and Zhitang Chen.
\newblock Causal discovery with reinforcement learning.
\newblock \emph{arXiv preprint arXiv:1906.04477}, 2019.

\bibitem[Sun et~al.(2007)Sun, Janzing, Sch{\"o}lkopf, and
  Fukumizu]{sun2007kernel}
Xiaohai Sun, Dominik Janzing, Bernhard Sch{\"o}lkopf, and Kenji Fukumizu.
\newblock A kernel-based causal learning algorithm.
\newblock In \emph{Proceedings of the 24th international conference on Machine
  learning}, pages 855--862. ACM, 2007.

\bibitem[Zhang et~al.(2012)Zhang, Peters, Janzing, and
  Sch{\"o}lkopf]{zhang2012kernel}
Kun Zhang, Jonas Peters, Dominik Janzing, and Bernhard Sch{\"o}lkopf.
\newblock Kernel-based conditional independence test and application in causal
  discovery.
\newblock \emph{arXiv preprint arXiv:1202.3775}, 2012.

\bibitem[Monti et~al.(2019)Monti, Zhang, and Hyvarinen]{monti2019causal}
Ricardo~Pio Monti, Kun Zhang, and Aapo Hyvarinen.
\newblock Causal discovery with general non-linear relationships using
  non-linear ica.
\newblock \emph{arXiv preprint arXiv:1904.09096}, 2019.

\bibitem[Eaton and Murphy(2007{\natexlab{c}})]{eatonuai}
Daniel Eaton and Kevin Murphy.
\newblock {Bayesian structure learning using dynamic programming and MCMC}.
\newblock In \emph{Uncertainty in Artificial Intelligence}, pages 101--108,
  2007{\natexlab{c}}.

\bibitem[Hoyer et~al.(2009)Hoyer, Janzing, Mooij, Peters, and
  Sch{\"o}lkopf]{hoyer2009nonlinear}
Patrik~O Hoyer, Dominik Janzing, Joris~M Mooij, Jonas Peters, and Bernhard
  Sch{\"o}lkopf.
\newblock Nonlinear causal discovery with additive noise models.
\newblock In \emph{Advances in neural information processing systems}, pages
  689--696, 2009.

\bibitem[Peters et~al.(2011)Peters, Mooij, Janzing, and
  Sch\"{o}lkopf]{Peters2011b}
J.~Peters, J.~M. Mooij, D.~Janzing, and B.~Sch\"{o}lkopf.
\newblock Identifiability of causal graphs using functional models.
\newblock In \emph{Proceedings of the 27th Annual Conference on {U}ncertainty
  in {A}rtificial {I}ntelligence ({UAI})}, pages 589--598, 2011.

\bibitem[Daniusis et~al.(2012)Daniusis, Janzing, Mooij, Zscheischler, Steudel,
  Zhang, and Sch{\"o}lkopf]{daniusis2012inferring}
Povilas Daniusis, Dominik Janzing, Joris Mooij, Jakob Zscheischler, Bastian
  Steudel, Kun Zhang, and Bernhard Sch{\"o}lkopf.
\newblock Inferring deterministic causal relations.
\newblock \emph{arXiv preprint arXiv:1203.3475}, 2012.

\bibitem[Budhathoki and Vreeken(2017)]{Budhathoki17}
Kailash Budhathoki and Jilles Vreeken.
\newblock Causal inference by stochastic complexity.
\newblock \emph{arXiv:1702.06776}, 2017.

\bibitem[Mitrovic et~al.(2018)Mitrovic, Sejdinovic, and
  Teh]{mitrovic2018causal}
Jovana Mitrovic, Dino Sejdinovic, and Yee~Whye Teh.
\newblock Causal inference via kernel deviance measures.
\newblock In \emph{Advances in Neural Information Processing Systems}, pages
  6986--6994, 2018.

\bibitem[Peters et~al.(2016)Peters, B{\"u}hlmann, and
  Meinshausen]{peters2016causal}
Jonas Peters, Peter B{\"u}hlmann, and Nicolai Meinshausen.
\newblock Causal inference by using invariant prediction: identification and
  confidence intervals.
\newblock \emph{Journal of the Royal Statistical Society: Series B (Statistical
  Methodology)}, 78\penalty0 (5):\penalty0 947--1012, 2016.

\bibitem[Ghassami et~al.(2017)Ghassami, Salehkaleybar, Kiyavash, and
  Zhang]{ghassami2017learning}
AmirEmad Ghassami, Saber Salehkaleybar, Negar Kiyavash, and Kun Zhang.
\newblock Learning causal structures using regression invariance.
\newblock In \emph{Advances in Neural Information Processing Systems}, pages
  3011--3021, 2017.

\bibitem[Rojas-Carulla et~al.(2018)Rojas-Carulla, Sch{\"o}lkopf, Turner, and
  Peters]{rojas2018invariant}
Mateo Rojas-Carulla, Bernhard Sch{\"o}lkopf, Richard Turner, and Jonas Peters.
\newblock Invariant models for causal transfer learning.
\newblock \emph{The Journal of Machine Learning Research}, 19\penalty0
  (1):\penalty0 1309--1342, 2018.

\bibitem[Bengio et~al.(2019)Bengio, Deleu, Rahaman, Ke, Lachapelle, Bilaniuk,
  Goyal, and Pal]{bengio2019meta}
Yoshua Bengio, Tristan Deleu, Nasim Rahaman, Rosemary Ke, S{\'e}bastien
  Lachapelle, Olexa Bilaniuk, Anirudh Goyal, and Christopher Pal.
\newblock A meta-transfer objective for learning to disentangle causal
  mechanisms.
\newblock \emph{arXiv preprint arXiv:1901.10912}, 2019.

\bibitem[Guyon(2013)]{guyoncause}
Isabelle Guyon.
\newblock Cause-effect pairs kaggle competition, 2013.
\newblock \emph{URL https://www. kaggle. com/c/cause-effect-pairs}, page 165,
  2013.

\bibitem[Guyon(2014)]{guyonchalearn}
Isabelle Guyon.
\newblock Chalearn fast causation coefficient challenge, 2014.
\newblock \emph{URL https://www. codalab. org/competitions/1381}, page 165,
  2014.

\bibitem[Lopez-Paz et~al.(2015)Lopez-Paz, Muandet, Sch{\"o}lkopf, and
  Tolstikhin]{lopez2015towards}
David Lopez-Paz, Krikamol Muandet, Bernhard Sch{\"o}lkopf, and Iliya
  Tolstikhin.
\newblock Towards a learning theory of cause-effect inference.
\newblock In \emph{International Conference on Machine Learning}, pages
  1452--1461, 2015.

\bibitem[Kalainathan et~al.(2018)Kalainathan, Goudet, Guyon, Lopez-Paz, and
  Sebag]{kalainathan2018sam}
Diviyan Kalainathan, Olivier Goudet, Isabelle Guyon, David Lopez-Paz, and
  Mich{\`e}le Sebag.
\newblock Sam: Structural agnostic model, causal discovery and penalized
  adversarial learning.
\newblock \emph{arXiv preprint arXiv:1803.04929}, 2018.

\bibitem[Goudet et~al.(2018)Goudet, Kalainathan, Caillou, Guyon, Lopez-Paz, and
  Sebag]{goudet2018learning}
Olivier Goudet, Diviyan Kalainathan, Philippe Caillou, Isabelle Guyon, David
  Lopez-Paz, and Michele Sebag.
\newblock Learning functional causal models with generative neural networks.
\newblock In \emph{Explainable and Interpretable Models in Computer Vision and
  Machine Learning}, pages 39--80. Springer, 2018.

\bibitem[Ivanov et~al.(2018)Ivanov, Figurnov, and
  Vetrov]{ivanov2018variational}
Oleg Ivanov, Michael Figurnov, and Dmitry Vetrov.
\newblock Variational autoencoder with arbitrary conditioning.
\newblock \emph{arXiv preprint arXiv:1806.02382}, 2018.

\bibitem[Li et~al.(2019)Li, Akbar, and Oliva]{li2019flow}
Yang Li, Shoaib Akbar, and Junier~B Oliva.
\newblock Flow models for arbitrary conditional likelihoods.
\newblock \emph{arXiv preprint arXiv=1909.06319}, 2019.

\bibitem[Yoon et~al.(2018)Yoon, Jordon, and Van Der~Schaar]{yoon2018gain}
Jinsung Yoon, James Jordon, and Mihaela Van Der~Schaar.
\newblock Gain: Missing data imputation using generative adversarial nets.
\newblock \emph{arXiv preprint arXiv:1806.02920}, 2018.

\bibitem[Douglas et~al.(2017)Douglas, Zarov, Gourgoulias, Lucas, Hart, Baker,
  Sahani, Perov, and Johri]{douglas2017universal}
Laura Douglas, Iliyan Zarov, Konstantinos Gourgoulias, Chris Lucas, Chris Hart,
  Adam Baker, Maneesh Sahani, Yura Perov, and Saurabh Johri.
\newblock A universal marginalizer for amortized inference in generative
  models.
\newblock \emph{arXiv preprint arXiv:1711.00695}, 2017.

\bibitem[Goodfellow et~al.(2014)Goodfellow, Pouget-Abadie, Mirza, Xu,
  Warde-Farley, Ozair, Courville, and Bengio]{goodfellow2014generative}
Ian Goodfellow, Jean Pouget-Abadie, Mehdi Mirza, Bing Xu, David Warde-Farley,
  Sherjil Ozair, Aaron Courville, and Yoshua Bengio.
\newblock Generative adversarial nets.
\newblock In \emph{Advances in neural information processing systems}, pages
  2672--2680, 2014.

\bibitem[Heinze-Deml et~al.(2018{\natexlab{a}})Heinze-Deml, Peters, and
  Meinshausen]{heinze2018invariant}
Christina Heinze-Deml, Jonas Peters, and Nicolai Meinshausen.
\newblock Invariant causal prediction for nonlinear models.
\newblock \emph{Journal of Causal Inference}, 6\penalty0 (2),
  2018{\natexlab{a}}.

\bibitem[Sch{\"o}lkopf et~al.(2012)Sch{\"o}lkopf, Janzing, Peters, Sgouritsa,
  Zhang, and Mooij]{SchJanPetSgoetal12}
Bernhard Sch{\"o}lkopf, Dominik Janzing, Jonas Peters, Eleni Sgouritsa, Kun
  Zhang, and Joris Mooij.
\newblock On causal and anticausal learning.
\newblock In J.~Langford and J.~Pineau, editors, \emph{Proceedings of the 29th
  International Conference on Machine Learning (ICML)}, pages 1255--1262, New
  York, NY, USA, 2012. Omnipress.

\bibitem[Beinlich et~al.(1989)Beinlich, Suermondt, Chavez, and
  Cooper]{beinlich1989alarm}
Ingo~A Beinlich, Henri~Jacques Suermondt, R~Martin Chavez, and Gregory~F
  Cooper.
\newblock The alarm monitoring system: A case study with two probabilistic
  inference techniques for belief networks.
\newblock In \emph{AIME 89}, pages 247--256. Springer, 1989.

\bibitem[Kristensen and Rasmussen(2002)]{kristensen2002use}
Kristian Kristensen and Ilse~A Rasmussen.
\newblock The use of a bayesian network in the design of a decision support
  system for growing malting barley without use of pesticides.
\newblock \emph{Computers and Electronics in Agriculture}, 33\penalty0
  (3):\penalty0 197--217, 2002.

\bibitem[Heinze-Deml et~al.(2018{\natexlab{b}})Heinze-Deml, Maathuis, and
  Meinshausen]{heinze2018causal}
Christina Heinze-Deml, Marloes~H Maathuis, and Nicolai Meinshausen.
\newblock Causal structure learning.
\newblock \emph{Annual Review of Statistics and Its Application}, 5:\penalty0
  371--391, 2018{\natexlab{b}}.

\bibitem[Kingma and Ba(2014)]{kingma2014adam}
Diederik~P Kingma and Jimmy Ba.
\newblock Adam: A method for stochastic optimization.
\newblock \emph{arXiv preprint arXiv:1412.6980}, 2014.

\end{thebibliography}

\clearpage

\section{Annexes}

\subsection{Training Algorithm} \label{algo}
Algorithm \ref{trainingloops} shows the pseudocode of the method described in \S\ref{method}. Typical values for the loop trip counts are found in \S\ref{appendix_samplecomplexity}.

\begin{algorithm*}[t]
\footnotesize
\caption{Training Algorithm}\label{trainingloops}
\begin{algorithmic}[1]
\Procedure{Training}{SCM Black Box Distribution $D$, with $M$ nodes and $N$ categories}
\State \textbf{Let} $i$ an index from $0$ to $M-1$
\Statex
\For{$I$ iterations, or until convergence,}
\For{$F$ functional parameter training steps}                   \Comment{\textbf{Phase 1}}
  \State $X \sim D$
  \State $C \sim \mathrm{Ber}(\sigma(\gamma))$
  \State $L = -\log P(X|C\,;\, \theta)$
  \State $\theta_{t+1} \gets \mathrm{Adam}(\theta_{t}, \nabla_\theta L)$
\EndFor
\Statex
\For{$Q$ interventions}
  \Statex                                                       \Comment{\textbf{Phase 2}}
  \State {\texttt{I\_N} $\gets \texttt{randint($0$, $M-1$)}$}               \Comment{Black Box, hidden}
  \State {$D_\textrm{int} := $ $D$ with intervention on node \texttt{I\_N}} \Comment{Black Box, hidden}
  \Statex
  \If{predicting intervention} \Comment{Phase 2 Prediction}
     \State $L_i \gets 0 \quad \forall i$
     \For{$N_P$ prediction steps}
       \State $X \sim D_\textrm{int}$
       \For{$C_P$ configurations}
         \State $C \sim \mathrm{Ber}(\sigma(\gamma))$
         \State{$L_i \gets L_i - \log P_{i}(X|C_i; \theta_\textrm{slow}) \;\forall i$}
       \EndFor
     \EndFor
     \State {\texttt{I\_N} $\gets \texttt{argmax}(L_i)$}
  \EndIf
  \Statex
  \State \texttt{gammagrads, logregrets = [], []}                           \Comment{Phase 2 Scoring}
  \For{$N_S$ scoring steps}
    \State $X \sim D_\textrm{int}$
    \State \texttt{gammagrad, logregret = 0, 0} 
    \For{$C_S$ configurations}
      \State $C \sim \mathrm{Ber}(\sigma(\gamma))$
      \State $L_i = -\log P_i(X|C_i; \theta_\textrm{slow}) \quad \forall i$
      \State $\texttt{gammagrad += } \sigma(\gamma) - C$                    \Comment{Collect $\sigmoid(\gamma) - C$ for Equation \ref{eq:metatransferloss}}
      \State $\texttt{logregret += } \sum\limits_{i \ne \texttt{I\_N}} L_i$ \Comment{Collect $\mathcal{L}_{C\makebox[0mm][l]{\tiny\raisebox{1.2mm}[0mm][0mm]{\hspace{-0.5mm}${}^{(k)}$}},i}\,(X)$ for Equation \ref{eq:metatransferloss}}
    \EndFor
    \State \texttt{gammagrads.append(gammagrad)}
    \State \texttt{logregrets.append(logregret)}
  \EndFor
  \Statex                                                                   \Comment{\textbf{Phase 3}}
  \State{$\displaystyle g_{ij} = \frac{\sum_{k} ( \sigma(\gamma_{ij}) - c^{(k)}_{ij} ) \mathcal{L}_{C\makebox[0mm][l]{\tiny\raisebox{1.2mm}[0mm][0mm]{\hspace{-0.5mm}${}^{(k)}$}},i}\,(X) }{\sum_k \mathcal{L}_{C\makebox[0mm][l]{\tiny\raisebox{1.2mm}[0mm][0mm]{\hspace{-0.5mm}${}^{(k)}$}},i}\,(X)}$} \Comment{Gradient Estimator, Equation \ref{eq:metatransferloss}}
  \State $\displaystyle g \gets g + \nabla_\gamma\left(\lambda_\textrm{sparse}\,L_\textrm{sparse}(\gamma) + \lambda_\textrm{DAG} \,L_\textrm{DAG} (\gamma)\right)$ \Comment{Regularizers}
  \State $\displaystyle \gamma_{t+1} \gets \mathrm{Adam}(\gamma_{t}, g)$
\EndFor
\EndFor
\EndProcedure
\end{algorithmic}
\end{algorithm*}

\subsection{Preliminaries}
\label{sec:appendix_prelim}

\paragraph{Interventions.}\label{appendix_intervention}
In a purely-observational setting, it is known that causal graphs can be distinguished only up to a Markov equivalence class. In order to identify the true causal graph intervention data is needed \citep{eberhardt2012number}. Several types of common \textit{interventions} may be available \citep{eaton2007exact}. These are:
\textit{No intervention:} only observational data is obtained from the ground truth causal model. \textit{Hard/perfect:} the value of a single or several variables is fixed and then ancestral sampling is performed on the other variables. \textit{Soft/imperfect:} the conditional distribution of the variable on which the intervention is performed is changed. \textit{Uncertain:} the learner is not sure of which variable exactly the intervention affected directly. 
Here we make use of soft interventions for several reasons: First, they include hard interventions as a limiting case and hence are more general. Second, in many real-world scenarios, it is more difficult to perform a hard intervention compared to a soft one.
We also deal with a special case of uncertain interventions, where the variable selected for intervention is random and unknown. We call these \emph{unidentified} or \emph{unknown} interventions. 

\paragraph{Intervention setup.}
For our experiments, the groundtruth models of the synthetic datasets are modeled by neural networks as described in section \ref{exp_setup_synthetic}. Each neural network  models the relationship of the causal parents and a variable. We perform our intervention by first randomly selecting which variable to intervene on, then soft-intervening on it. The selected variable is sampled from a uniform distribution. The soft intervention is a reinitialization of its neural network's parameters.

\paragraph{Causal sufficiency.}
The inability to distinguish which causal graph, within a Markov equivalence class, is the correct one in the purely-observational setting is called the \textit{identifiability problem}. In our setting, all variables are observed (there are no latent confounders) and all interventions are random and independent. Hence, within our setting, if the interventions are known, then the true causal graph is always identifiable in principle \citep{eberhardt2012number,heinze2018causal}.  We also consider here situations where a single variable is randomly selected and intervened upon with a soft or imprecise intervention, its identity is unknown and must be inferred. In this case, there is no theoretical guarantee that the causal graph is identifiable. However, there is existing work \citet{peters2016causal} that handles this scenario and the proposed method is also proven to work empirically. 

\paragraph{Faithfulness.}
It is possible for causally-related variables to be probabilitistically independent purely by happenstance, such as when causal effects along multiple paths cancel out. This is called \textit{unfaithfulness}. We assume that \textit{faithfulness} holds, since the $\gamma$ gradient estimate is extracted from shifts in probability distributions. However, because of the ``soft'' nature of our interventions and their infinite variety, it would be exceedingly unlikely for cancellation-related unfaithfulness to persist throughout the causal-learning procedure.

\subsection{Experimental setup }\label{exp_setup}

For all datasets, the weight parameters for the learned model is initialized randomly. In order to not bias the structural parameters, all $\sigma(\gamma)$ are initialized to $0.5$ in the beginning of training. Details of hyperparameters of the learner model are described in Section \ref{appendix_hyperparam}. The experimental setup for the groundtruth model for the synthetic data can be found in Section \ref{exp_setup_synthetic} and the details for the real world data are described in Section \ref{exp_setup_bnlearn}.

\subsection{Model setup}\label{appendix_modelsetup}

As discussed in section \ref{method}, we model the $M$ variables in the graph using $M$ independent MLPs,  each possesses an input layer of $M \times N$ neurons (for $M$ one-hot vectors of length $N$ each), a single hidden layer chosen arbitrarily to have $\max(4M,4N)$ neurons with a LeakyReLU activation of slope $0.1$, and a linear output layer of $N$ neurons representing the unnormalized log-probabilities of each category (a softmax then recovers the conditional probabilities from these logits). To force $f_i$ to rely exclusively on the direct ancestor set $pa(i,C)$ under adjacency matrix $C$ (See Eqn.~\ref{eq:metatransferloss}), the one-hot input vector $X_j$ for variable $X_i$'s MLP is masked by the Boolean element $c_{ij}$. The functional parameters of the MLP are the set $\theta = \{\texttt{W0}_{ihjn}, \texttt{B0}_{ih}, \texttt{W1}_{inh}, \texttt{B1}_{in}\}$.An example of the multi-MLP architecture with $M$=3 categorical variables of $N$=2 categories is shown in Figure \ref{model_figure}.
\subsection{Hyperparameters} \label{appendix_hyperparam}

\paragraph{Learner model.} All experiments on the synthetic graphs of size 3-8 use the same hyperparameters. Both the functional and structural parameters are optimized  using the Adam optimizer \cite{kingma2014adam}. We use a learning rate of $5e-2$  with alpha of $0.9$ for the functional parameters, and we use a learning rate of $5e-3$ with alpha of $0.1$ for the structural parameters. We perform $5$ runs of each experiment with random seeds $1-5$ and error bars are plotted for various graphs from size $3$ to $8$ in Figure \ref{fig:synthetic_data_results}. We use a batch size of $256$. The L1 norm regularizer is set to $0.1$ and the $DAG$ regularizer is set to $0.5$ for all experiments. For each $\gamma$ update step, we sample $25$ structural configurations from the current $\gamma$. In all experiments, we use $100$ batches from the interventional distribution to predict the intervened node.

\subsection{Synthetic data}\label{exp_setup_synthetic}
\begin{figure}[h]
    \centering
    \begin{tabular}{l}
    \includegraphics[scale=0.7]{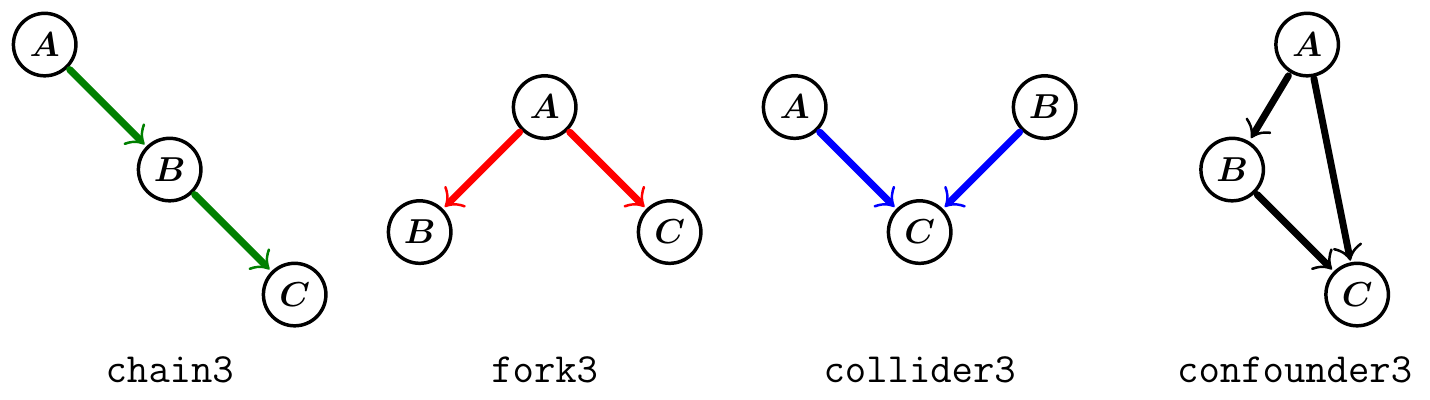}\\
    \end{tabular}
    \caption{Every possible 3-variable connected DAG.
    }
    \label{fig:3varCE}
\end{figure}

\paragraph{Synthetic datasets.} The synthetic datasets in the paper are modeled by neural networks. All neural networks are 2 layered feed forward neural networks (MLPs) with Leaky ReLU activations between layers. The parameters of the neural network are initialized orthogonally within the range of $(-2.5, 2.5)$. This range was selected such that they output a non-trivial distribution. The biases are initialized uniformly between $(-1.1, 1.1)$.

SCM with $n$ variables are modeled by $n$ feedforward neural networks (MLPs) as described in \S\ref{sec:modeldescription}. We assume an acyclic causal graph so that we may easily sample from them. Hence, given any pair of random variables $A$ and $B$, either $A \xrightarrow{} B$, $B \xrightarrow{ }A$ or $A$ and $B$ are independent.

\begin{figure}
    \centering
    \subfigure[\textbf{chainN}]{\includegraphics[ width=0.32\textwidth]{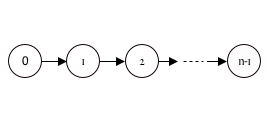}}
    \subfigure[\textbf{colliderN}]{\includegraphics[ width=0.32\textwidth]{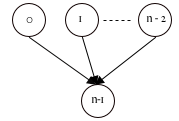}} 
    \subfigure[\textbf{bidiagN}]{\includegraphics[ width=0.32\textwidth]{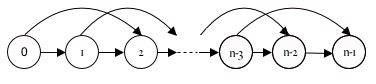}}\par
    \subfigure[\textbf{fullN}]{\includegraphics[ width=0.32\textwidth]{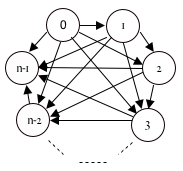}}
    \subfigure[\textbf{jungleN}]{\includegraphics[ width=0.32\textwidth]{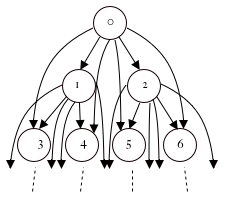}}
    \caption{Figures for various synthetic graphs. \texttt{chain},  \texttt{collider}, \texttt{bidiagonal}, \texttt{full} and \texttt{jungle} graph.}
    \label{fig:synthetic_graphs}
\end{figure}

The MLP representing the ground-truth SCM has its weights $\theta$ initialized use orthogonal initialization with gain $2.5$ and the biases are initialized using a uniform initialization between $-1.1$ and $1.1$, which was empirically found to yield "interesting" yet learnable random SCMs.

We study a variety of SCMs with different ground-truth edge structures $\gamma$. Our selection of synthetic graphs explores various extremes in the space of DAGs, stress-testing \gls{SDI}. The \texttt{chain} graphs are the sparsest connected graphs possible, and are relatively easy to learn. The \texttt{bidiag} graphs are extensions of \texttt{chain} where there are 2-hops as well as single hops between nodes, doubling the number of edges and creating a meshed chain of forks and colliders. The \texttt{jungle} graphs are binary-tree-like graphs, but with each node connected directly to its grandparent in the tree as well. Half the nodes in a \texttt{jungle} graph are leaves, and the out-degree is up to 6. The \texttt{collider} graphs deliberately collide independent $M-1$ ancestors into the last node; They stress maximum in-degree. Lastly, the \texttt{full} graphs are the maximally dense DAGs. All nodes are direct parents of all nodes below them in the topological order. The maximum in- and out-degree are both $M-1$. These graphs are depicted in Figure \ref{fig:synthetic_graphs}.

\subsubsection{Synthetic data results} \label{appendix_synthetic_results}
The model can recover correctly all synthetic graphs with 10 variables or less, as shown in Figure \ref{fig:ce_all_graphs} and Table \ref{table:all_baseline_hamming}. For graphs larger than 10 variables, the model found it more challenging to recover the denser graphs (e.g. \texttt{fullM}), as shown in Table \ref{table:all_baseline_hamming}. Plots of the training curves showing average cross entropy (CE) and Area-Under-Curve(AUC/AUCROC) for edge probabilities of the learned graph against the ground-truth graph for synthetic SCMs with 3-13 variables are available in Figure \ref{fig:ce_all_graphs}.

\subsection{BnLearn data repository}\label{exp_setup_bnlearn}
\begin{figure}
    \centering
    \includegraphics[scale=0.30]{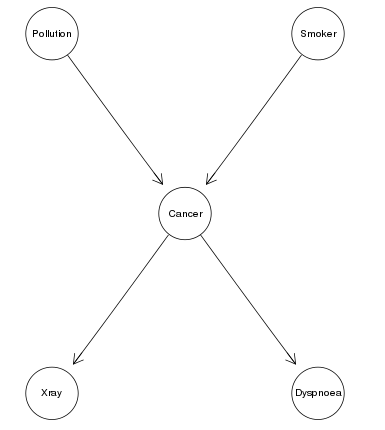}
    \includegraphics[scale=0.30]{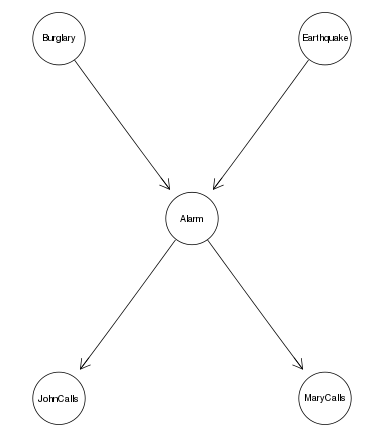}
    \includegraphics[scale=0.35]{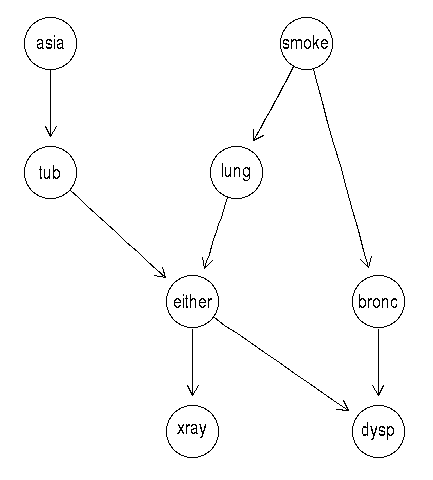}
    \caption{Left to right: Ground Truth SCM for Cancer, Groundtruth SCM for Earthquake, Groundtruth SCM for Asia.}
    \label{fig:earthquake_graph}
\end{figure}

The repo contains many datasets  with various sizes and structures modeling different variables. We evaluate the proposed method on 3 of the datasets in the repo, namely the Earthquake \citep{korb2010bayesian}, Cancer \citep{korb2010bayesian} and Asia \citep{lauritzen1988local} datasets. The ground-truth model structure for the Cancer \citep{korb2010bayesian} and Earthquake \citep{korb2010bayesian} datasets are shown in Figure \ref{fig:earthquake_graph}. Note that even though the structure for the 2 datasets seems to be the same, the conditional probability tables (CPTs) for these 2 datasets are very different and hence results in different structured causal models (SCMs) for the 2 datasets.

\begin{table*}[h]
 \centering  

\begin{tabular}{lrrrrrr}
 \toprule
{\bf \small Method}                      & \texttt{Asia} &   \texttt{chain8}   & \texttt{jungle8} & \texttt{collider7} & \texttt{collider8}& \texttt{full8} \\
 \midrule
{\bf \small \citep{zheng2018dags}}       &  14  &     24               &  14    &      11 &    18     &       21      \\
{\bf \small \citep{yu2019dag}}           &  10  &    7                &  12     &     6 &   7      &       25      \\
{\bf \small \citep{heinze2018invariant} } &   8  &   7                &        12         &    6      &  7   &    28           \\
{\bf \small \citep{peters2016causal}}    &   5  & 3                   &        8          &    4     &    2  &   16            \\
{\bf \small \citep{eaton2007belief}}      &   0  &         0         &        0         &       7  &    7 &        1      \\
{\bf \small \gls{SDI}s}                  &   0  &         0          &        0         &       0  &  0    &        0      \\
 \bottomrule
 \end{tabular}
 \caption{\textbf{Baseline  comparisons:} Hamming distance (lower is better) for learned and ground-truth edges on various graphs from both synthetic and real datasets, compared to \citep{peters2016causal}, \citep{heinze2018invariant}, \citep{eaton2007exact}, \citep{yu2019dag} and \citep{zheng2018dags}. The proposed \gls{SDI} is run on random seeds $1-5$ and we pick the worst performing model out of the random seeds in the table.} \label{table:all_baseline_hamming_supp}
\end{table*}

\begin{figure}[htb!]
    \centering
    \begin{subfigure}
    \centering
    \includegraphics[scale=0.6]{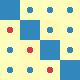}
    \includegraphics[scale=0.6]{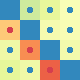}
    \includegraphics[scale=0.6]{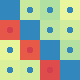}
    \label{fig::gamma::chain4}
    \end{subfigure}\hfill
    \begin{subfigure}
    \centering
    \includegraphics[scale=0.6]{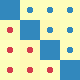}
    \includegraphics[scale=0.6]{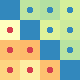}
    \includegraphics[scale=0.6]{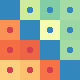}
    \label{fig::gamma::full4}
    \end{subfigure}
    \caption{Learned edges at three different stages of training. \textbf{Left}: \texttt{chain4} (chain graph with 4 variables). \textbf{Right}: \texttt{full4} (tournament graph with 4 variables).
    }\label{fig:gamma:synthetic_results}
\vspace{-4mm}
\end{figure}

Given that some of the CPTs contain very unlikely events, we have found it necessary to add a \textit{temperature} parameter in order to make them more frequent. The near-ground-truth MLP model's logit outputs are divided by the temperature before being used for sampling. Temperatures above 1 result in more uniform distributions for all causal variables; Temperatures below 1 result in less uniform, sharper distributions that peak around the most likely value. We find empirically that a temperaure of about 2 is required for our BnLearn benchmarks.

\vspace*{-1mm}
\subsection{Comparisons to other methods}\label{annex:comparison}
\vspace*{-1mm}

As described in section \ref{baseline_compare}, we compare to 5 other methods. The full comparison between \gls{SDI}s and other methods on various graphs can be found in Table \ref{table:all_baseline_hamming}.

One of these methods, DAG-GNN~\citet{yu2019dag}, outputs 3 graphs based on different criteria: best mean square error (MSE), best negative loglikelihood (NLL) and best evidence lower bound (ELBO). We report performance of all outputs of DAG-GNN~\citet{yu2019dag} in Table \ref{table:baseline_daggnn}, and the best one is selected for Table \ref{table:all_baseline_hamming}.

\begin{figure}[]
    \centering
    \begin{subfigure}
    \centering
    \includegraphics[scale=0.5]{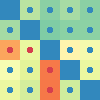}
    \includegraphics[scale=0.5]{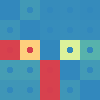}
    \includegraphics[scale=0.5]{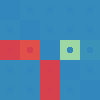}
    \end{subfigure}\hfill
     \begin{subfigure}
    \centering
     \includegraphics[scale=0.42]{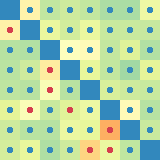}
    \includegraphics[scale=0.42]{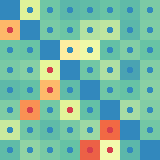}
    \includegraphics[scale=0.42]{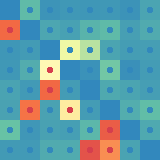}
    \end{subfigure}
    \caption{\textbf{Left:} Earthquake: Learned edges at three different stages of training. \textbf{Right:} Asia: Learned edges at three different stages of training.}
    \label{fig:gamma:earthquake}
    \vspace{-4mm}
\end{figure}

\begin{table}[h]
\centering  
{\begin{tabular}{lccccc}
\toprule
&{\bf \gls{SDI}}  & \small{Best MSE} & \small{Best NLL} & \small{Best Elbo} \\
\midrule
\texttt{Asia} & 0 & 10  &  10  & 13 \\
\texttt{chain8} & 0 & 7  &  7  & 7 \\
\texttt{jungle8} & 0 & 12  &  12  & 13 \\
\texttt{collider7} & 0 & 6  &  6  & 6 \\
\texttt{collider8} & 0 & 8  &  8  & 7 \\
\texttt{full8} & 0 & 27  &  25  & 27 \\
\bottomrule
\end{tabular} \\
}

\caption{\textbf{Baseline  comparisons:} Hamming distance (lower is better) for learned and ground-truth edges on \texttt{Asia} and various synthetic graphs. compared to DAG-GNN~\citet{yu2019dag}. DAG-GNN outputs 3 graphs according to different criterion. We show results on all outputs in this table and we show the best performing result in Table \ref{table:all_baseline_hamming}.}\label{table:baseline_daggnn}

\end{table}

\begin{table*}[h]
\centering
\setlength{\tabcolsep}{3pt}
\begin{tabular}[]{|ll|c|c|c|c|c|c|c|c|c|c|c|c|c|c|c|c|c|c|c|c|c|}
  \cline{3-20}
  \mc{1}{l}{}&&\mc{6}{|c|}{\texttt{Chain}}&\mc{6}{c|}{\texttt{Jungle}}&\mc{6}{c|}{\texttt{Full} }\\
  \cline{3-20}
  \mc{1}{l}{}&& \mc{1}{c}{3} & \mc{1}{c}{4} & \mc{1}{c}{5} & \mc{1}{c}{6} & \mc{1}{c}{7} & \mc{1}{c|}{8} 
              & \mc{1}{c}{3} & \mc{1}{c}{4} & \mc{1}{c}{5} & \mc{1}{c}{6} & \mc{1}{c}{7} & \mc{1}{c|}{8}  
              & \mc{1}{c}{3} & \mc{1}{c}{4} & \mc{1}{c}{5} & \mc{1}{c}{6} & \mc{1}{c}{7} & \mc{1}{c|}{8} \\
  \hline
    & \textbf{ldag=$0.5$, lsparse=$0.1$}  & 0 & 0 & 0 & 0 & 0 & 0 & 0 & 0 & 0 & 0 & 0 & 0 & 0 & 0 & 0 & 0 & 0 & 0\\
  \cline{3-20}
  & \textit{ldag=$0.5$, lsparse=$0.$}   & 0 & 0 & 0 & 0 & 0 & 0 & 0 & 0 & 0 & 0 & 0 & 0& 0 & 0 & 0 & 0 & 0 & 0\\
  \cline{3-20}
    & \textit{ldag=$0.$, lsparse=$0.1$} & 0 & 0 & 0 & \textbf{1} &0 & 0 & 0 & 0 & 0 & \textbf{1} & \textbf{3} & \textbf{1} &0 & 0 & 0 & 0 & \textbf{1} & \textbf{6} \\
  \hline
\end{tabular}\caption{\textbf{Regularizer:} \gls{SDI} performance measured by Hamming distance to the ground-truth graph. Comparisons are between \gls{SDI}s with different regularizer settings for different graphs. Our default setting is $\mathrm{ldag}=0.5$, $\mathrm{lsparse}=0.1$, with $\mathrm{ldag}$ the DAG regularization strength and $\mathrm{lsparse}$ the sparsity regularization strength. As shown in the table, \gls{SDI}s is not very sensitive to different regularizer settings. Tasks with non-zero Hamming distance (errors) are in bold.}\label{table:regularizer_comparison}
\end{table*}

\subsection{Sparsity of Ground-Truth Graph}\label{appendix_sparsity}
We evaluated the performance of \gls{SDI} on graphs of various size and sparsity to better understand the performance of the model. We evaluated the proposed model on 4 representative types of graphs in increasing order of density. They are the \texttt{chain}, \texttt{jungle},  \texttt{bidiag} and \texttt{full} graphs. As shown in the results in figure \ref{fig:sparsity}, for graphs of size 5 or smaller, there is almost no difference in the final results in terms of variance and sample complexity. However, as the graphs gets larger (than 6), the denser graphs (\texttt{full} graphs) gets progressively more difficult to learn compared to the sparser graphs (\texttt{chain}, \texttt{jungle} and \texttt{bidiag}). The models learned for denser graphs have higher complexity, higher variance and slightly worse results. 
\begin{figure}[!h]
    \centering
    \includegraphics[scale=0.65]{asy/chainN-gamma-ce-errorbars-appendix-vector.pdf}
    \includegraphics[scale=0.65]{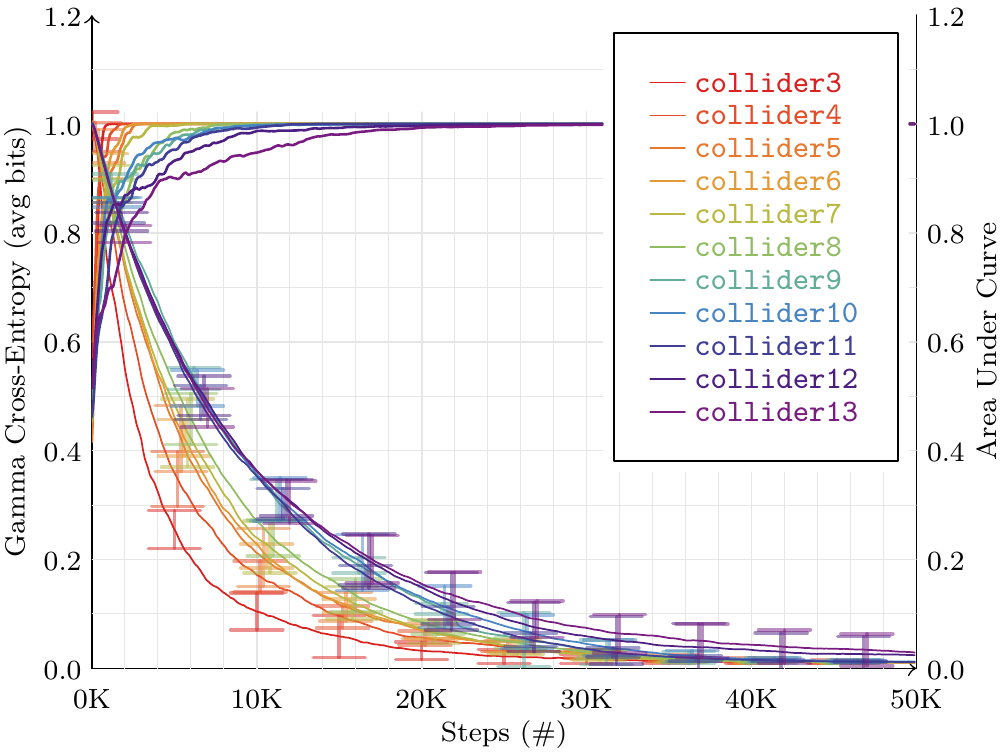}\\
    \includegraphics[scale=0.65]{asy/jungleN-gamma-ce-errorbars-appendix-vector.pdf}
    \includegraphics[scale=0.65]{asy/fullN-gamma-ce-errorbars-appendix-vector.pdf}
    \caption{Cross entropy (CE) and Area-Under-Curve (AUC/AUROC) for edge probabilities of learned graph against ground-truth for synthetic SCMs. Error bars represent $\pm 1\sigma$ over PRNG seeds 1-5. \textbf{Left to right, up to down}: 
    \texttt{chainM},\texttt{jungleM},\texttt{fullM},$M=3\ldots8$ ($9\ldots 13$ in Appendix \ref{appendix_synthetic_results}). Graphs (3-13 variables) all learn perfectly with AUROC reaching 1.0. However, denser graphs (fullM) take longer to converge.} \label{fig:ce_all_graphs}
\end{figure}

\subsection{Predicting interventions}
 \begin{figure}[!h]
    \centering
    \includegraphics[scale=0.45]{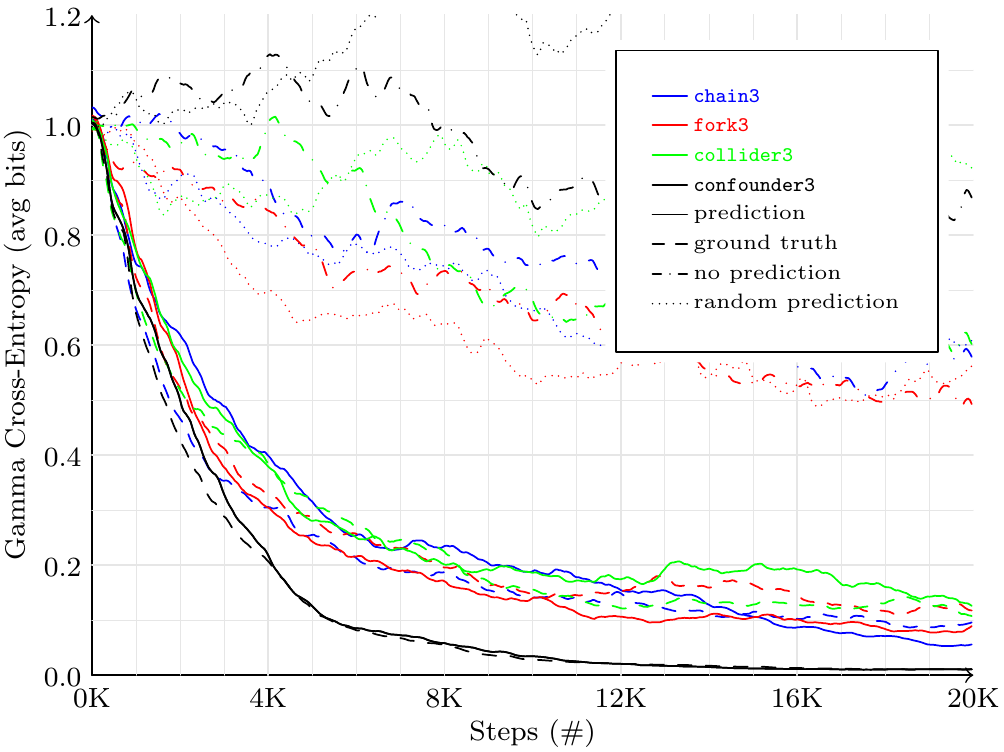}
    \includegraphics[scale=0.45]{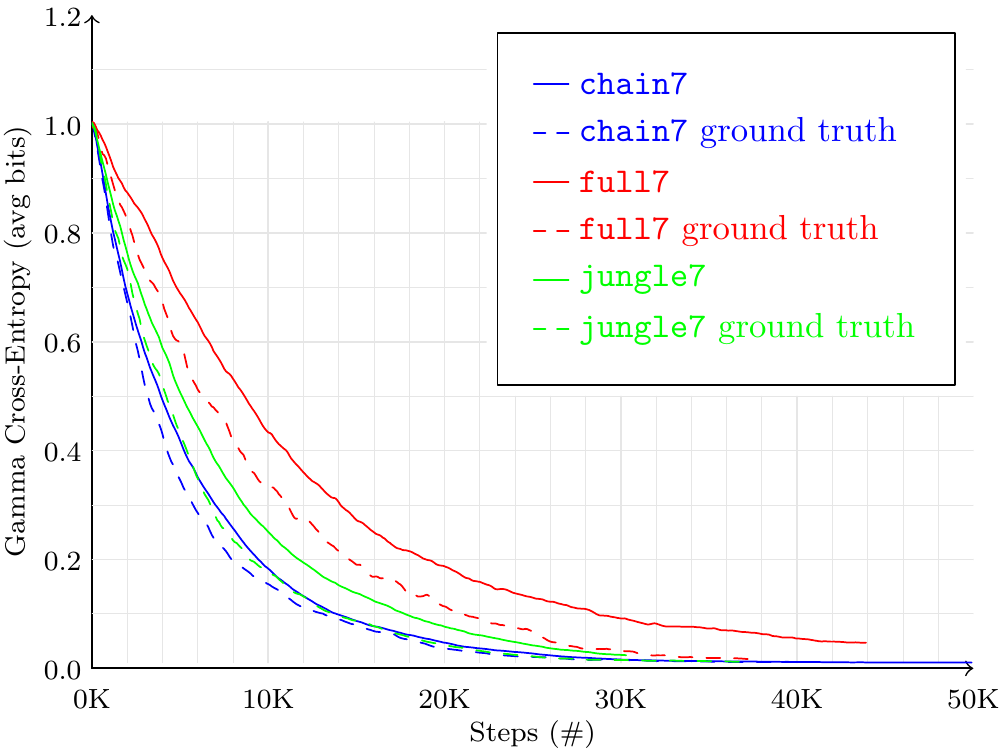}
    \caption{\textbf{Ablation Study of Intervention Prediction} Cross-entropy loss over time on multiple graphs and intervention prediction modes. \textbf{Left}: All 3-variable graphs. Solid/dashed lines: Ground-truth \& Prediction strategies. Dotted lines: Random- \& No-Prediction strategies. Training with prediction closely tracks ground-truth. \textbf{Right}: Comparison for 7-variable graphs, ground-truth against prediction strategy. Training with prediction still closely tracks ground-truth at larger scales.}
    \label{fig:predict-or-not}
\end{figure}

In Phase 2, we score graph configurations based on how well they fit the interventional data. We find that it is necessary to avoid disturbing the learned parameters of intervened variables, and to ignore its contribution to the total negative log-likelihood of the sample. Intuitively, this is because, having been intervened upon, that variable should be taken as a given. It should especially not be interpreted as a poorly-learned variable requiring a tuning of its functional parameters, because those functional parameters were not responsible for the value of that variable; The extrinsic intervention was.

Since an intervened variable is likely to be unusually poorly predicted, we heuristically determine that the most poorly predicted variable is the intervention variable. We then zero out its contribution to the log-likelihood of the sample and block gradient into its functional parameters.

Figure \ref{fig:predict-or-not} illustrates the necessity of this process. When using the prediction heuristic, the training curve closely tracks training with ground-truth knowledge of the identity of the intervention. If no prediction is made, or a random prediction is made, training proceeds much more slowly, or fails entirely.

\subsection{Sample complexity}\label{appendix_samplecomplexity}

Our method is heavily reliant on sampling of configurations and data in Phases 1 and 2. We present here the breakdown of the sample complexity. Let
\begin{itemize}
    \item $I$ be the number of iterations of the method, \hfill\textit{\quad\small(typical: 500-2000)}
    \item $B$ the number of samples per batch, \hfill\textit{\quad\small(typical: 256)}
    \item $F$ the number of functional parameter training iterations in Phase 1, \hfill\textit{\quad\small(typical: 10000)}
    \item $Q$ the number of interventions performed in Phase 2, \hfill\textit{\quad\small(typical: 100)}
    \item $N_P$ the number of data batches for prediction, \hfill\textit{\quad\small(typical: 100)}
    \item $C_P$ the number of graph configurations drawn per prediction data batch, \hfill\textit{\quad\small(typical: 10)}
    \item $N_S$ the number of data batches for scoring, \hfill\textit{\quad\small(typical: 10)}
    \item $C_S$ the number of graph configurations drawn per scoring data batch. \hfill\textit{\quad\small(typical: 20-30)}
\end{itemize}
Then the total number of interventions performed, and configurations and samples drawn, over an entire run are:
\begin{align}
    \textrm{Interventions} &= IQ = \gamma\textrm{ updates}\\
    \textrm{Samples} &= I(\underbrace{F}_\textrm{Phase 1} + \underbrace{Q(N_P + N_S)}_\textrm{Phase 2})B \\
    \textrm{Configurations} &= I(\underbrace{F}_\textrm{Phase 1} + \underbrace{Q(C_P N_P + C_S N_S)}_\textrm{Phase 2})
\end{align}
Because of the multiplicative effect of these factors, the number of data samples required can quickly spiral out of control. For typical values, as many as $500 \times 10000 \times 256 = 1.28\mathrm{e}9$ observational and $500 \times 100 \times (100+10) \times 256 = 1.408\mathrm{e}9$ interventional samples are required. To alleviate this problem slightly, we limit the number of samples generated for each intervention; This limit is usually 500-2000.

\begin{figure}[!h]
    \centering
    \includegraphics[scale=0.42]{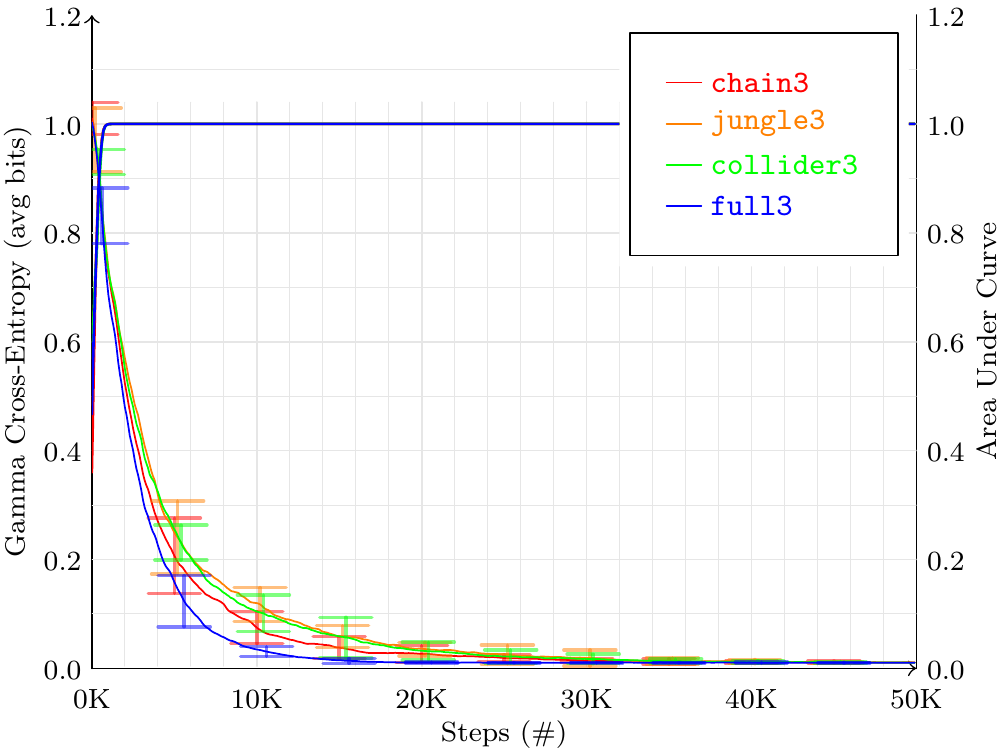}
    \includegraphics[scale=0.42]{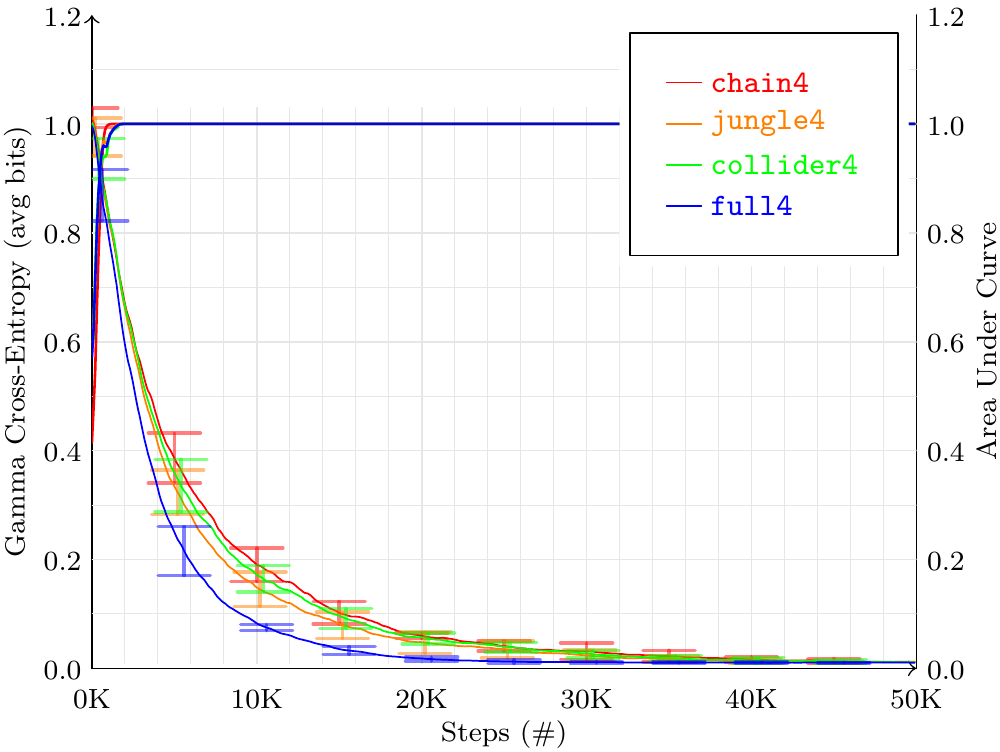}
    \includegraphics[scale=0.42]{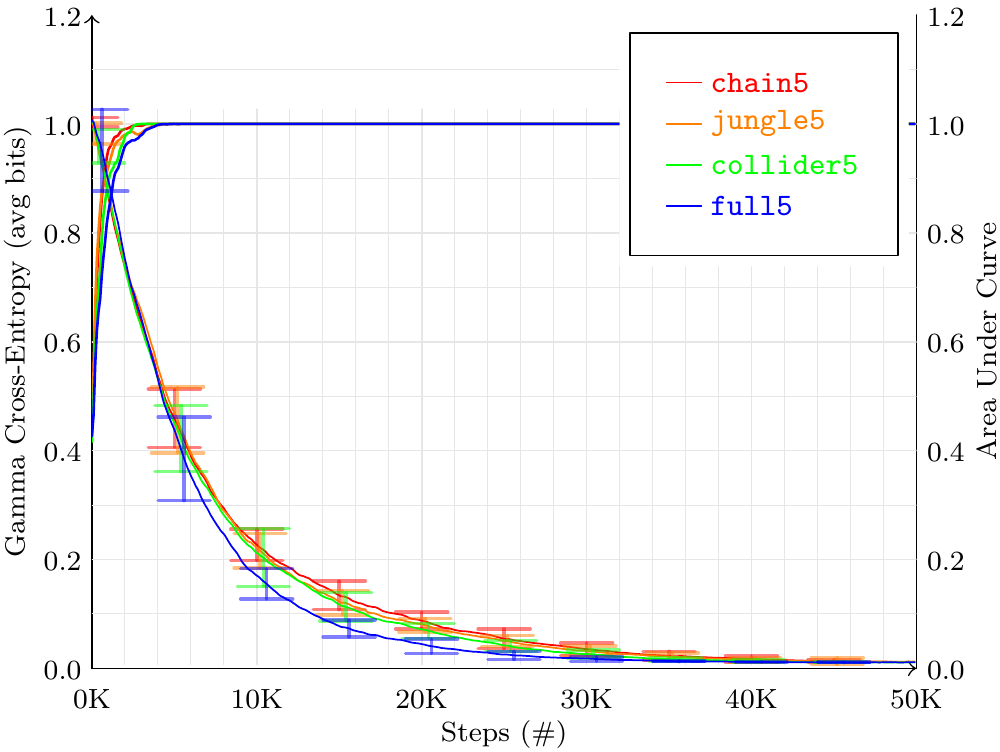}\\
    \includegraphics[scale=0.42]{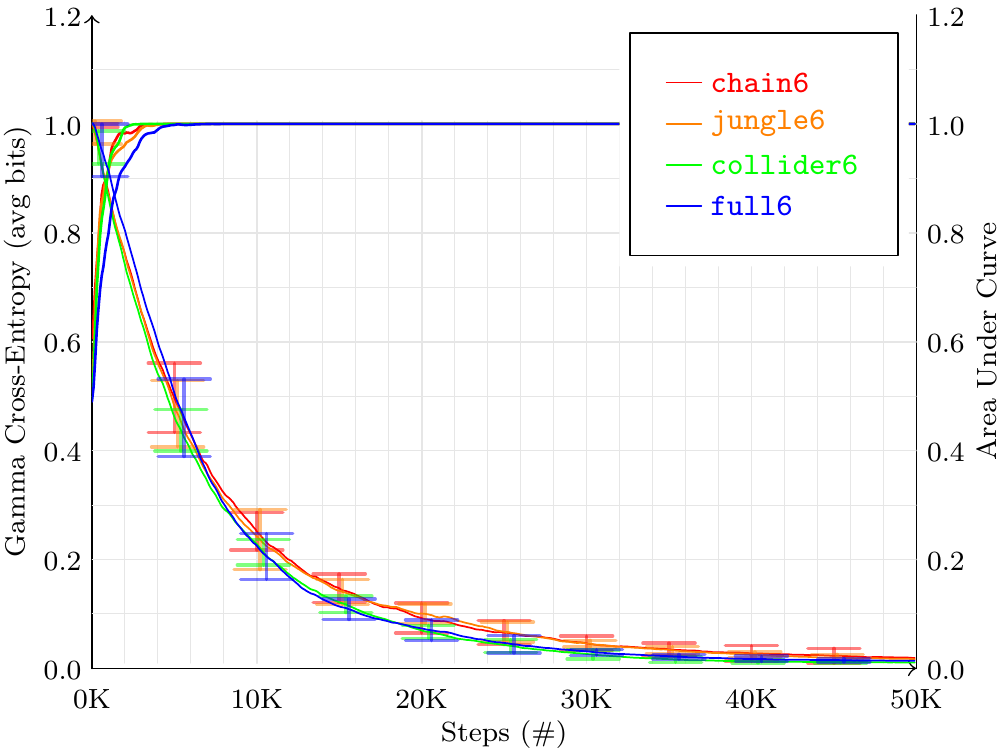}
    \includegraphics[scale=0.42]{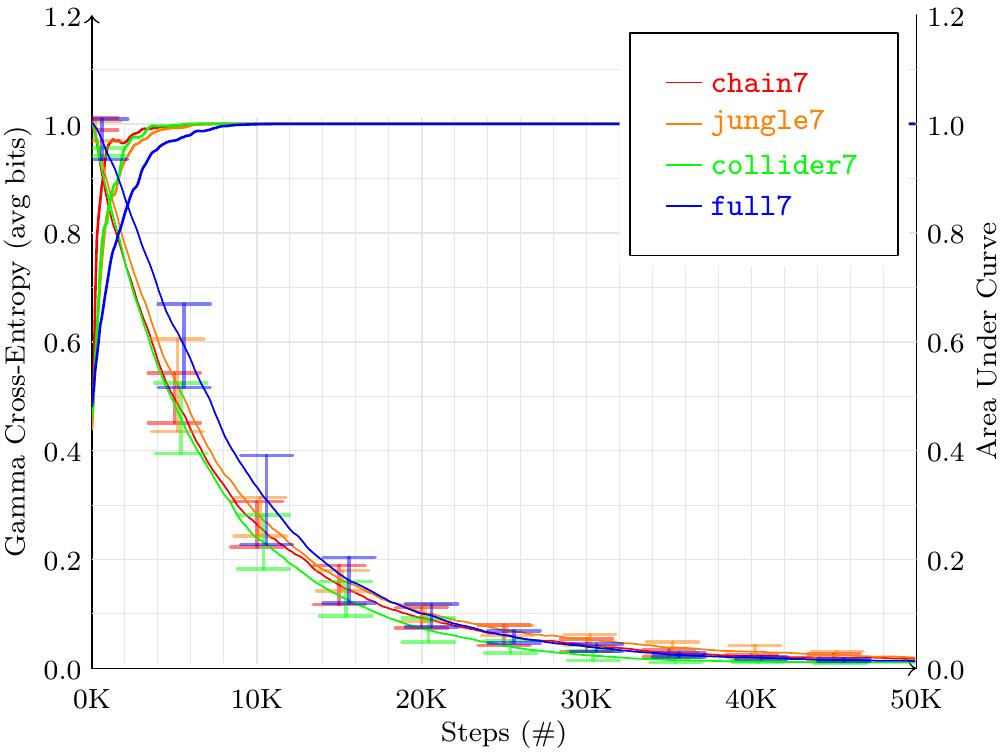}
    \includegraphics[scale=0.42]{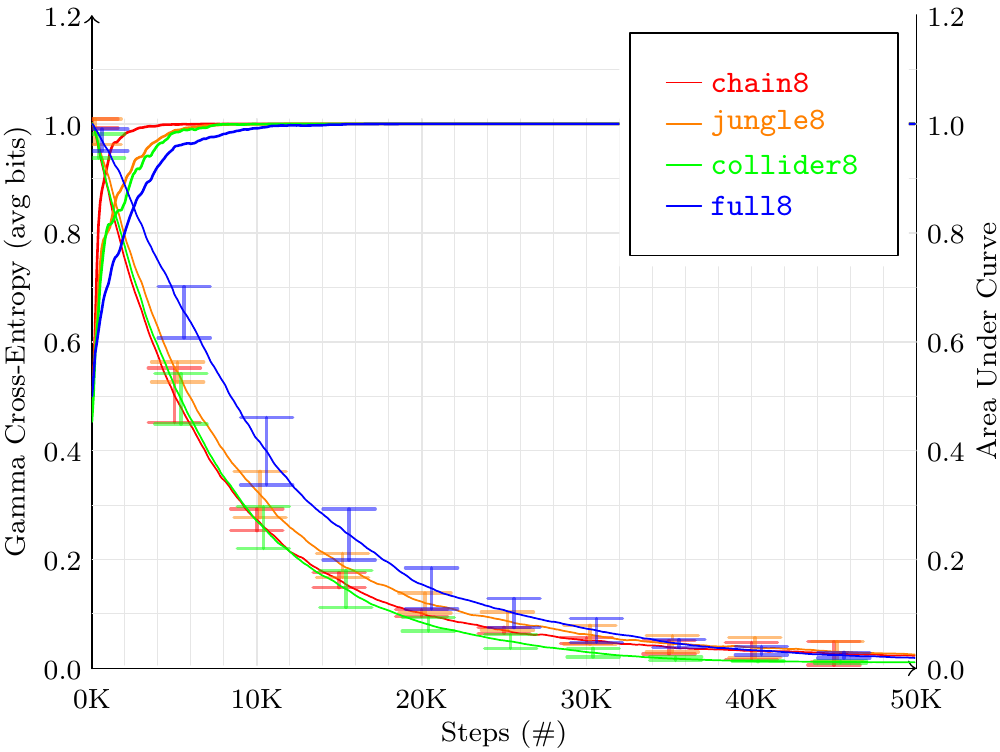}\\
    \includegraphics[scale=0.42]{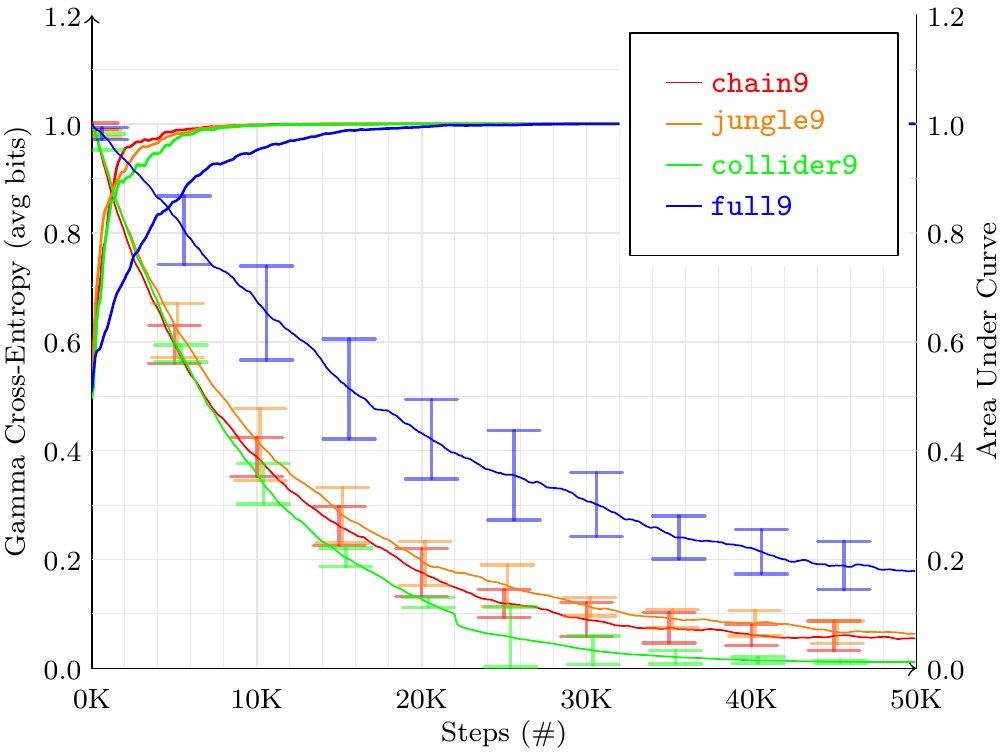}
    \includegraphics[scale=0.42]{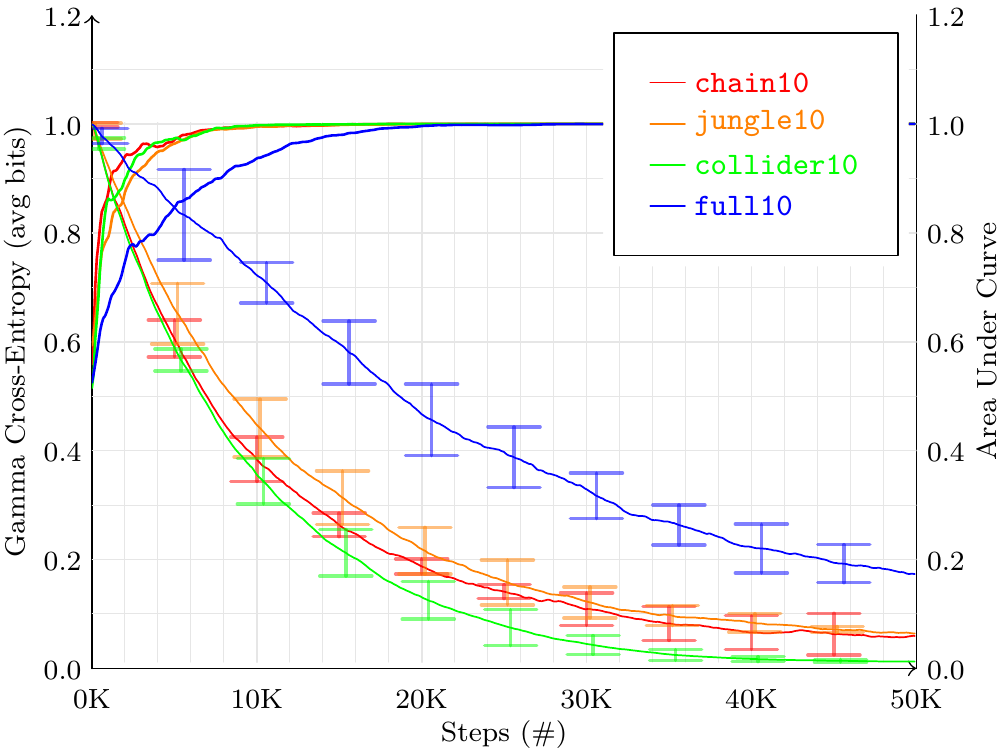}
    \includegraphics[scale=0.42]{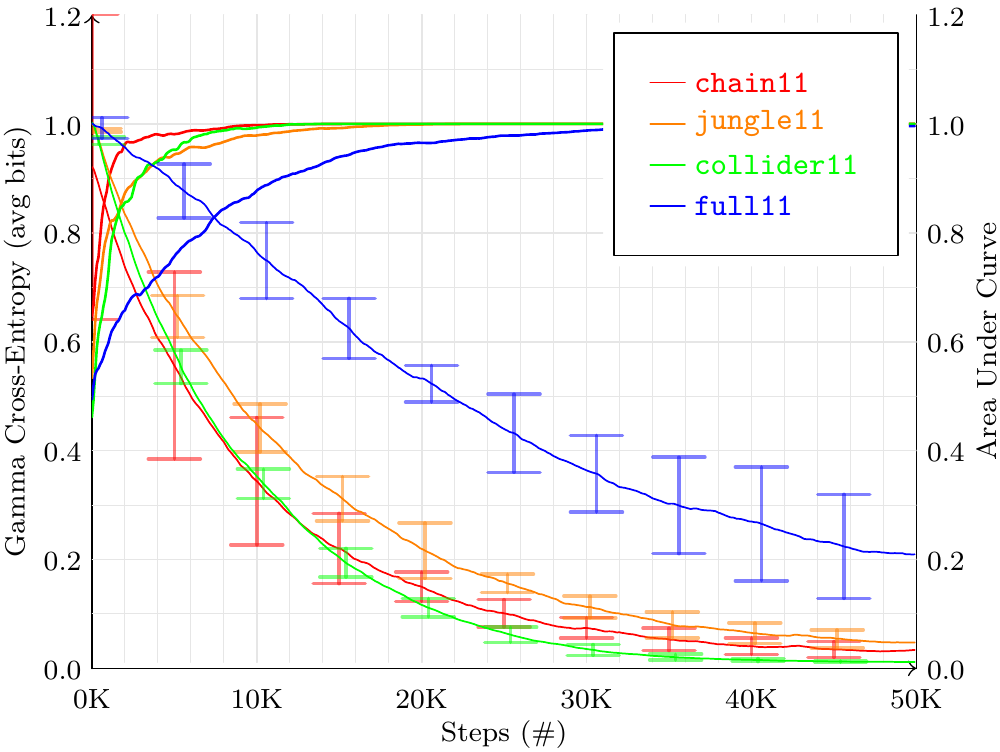}\\
    \includegraphics[scale=0.42]{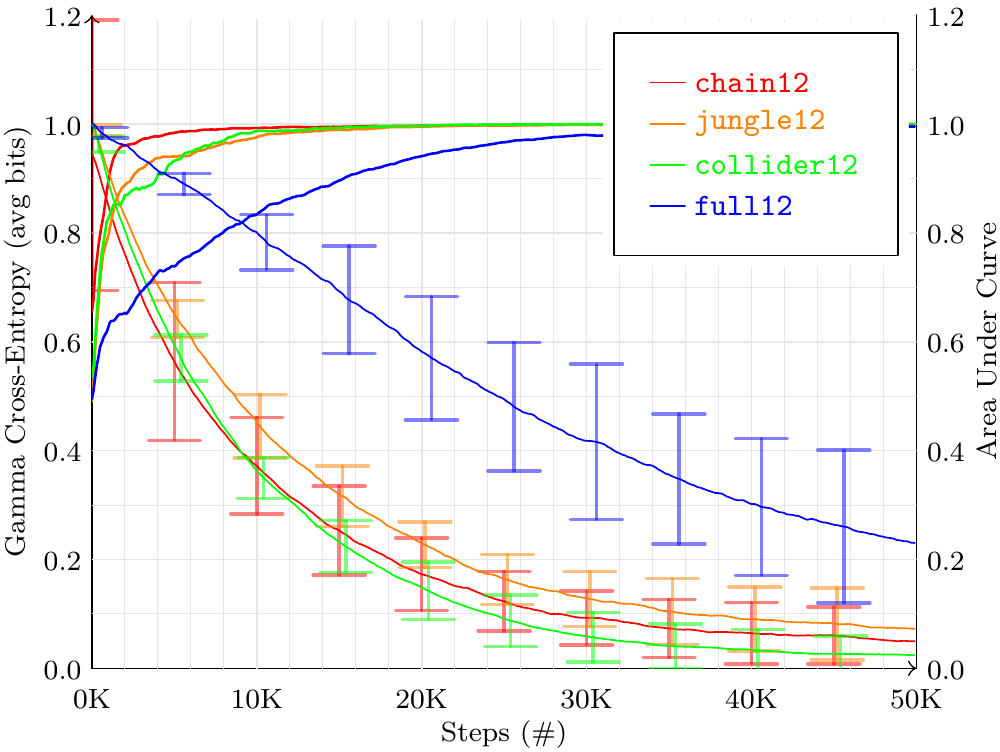}
    \includegraphics[scale=0.42]{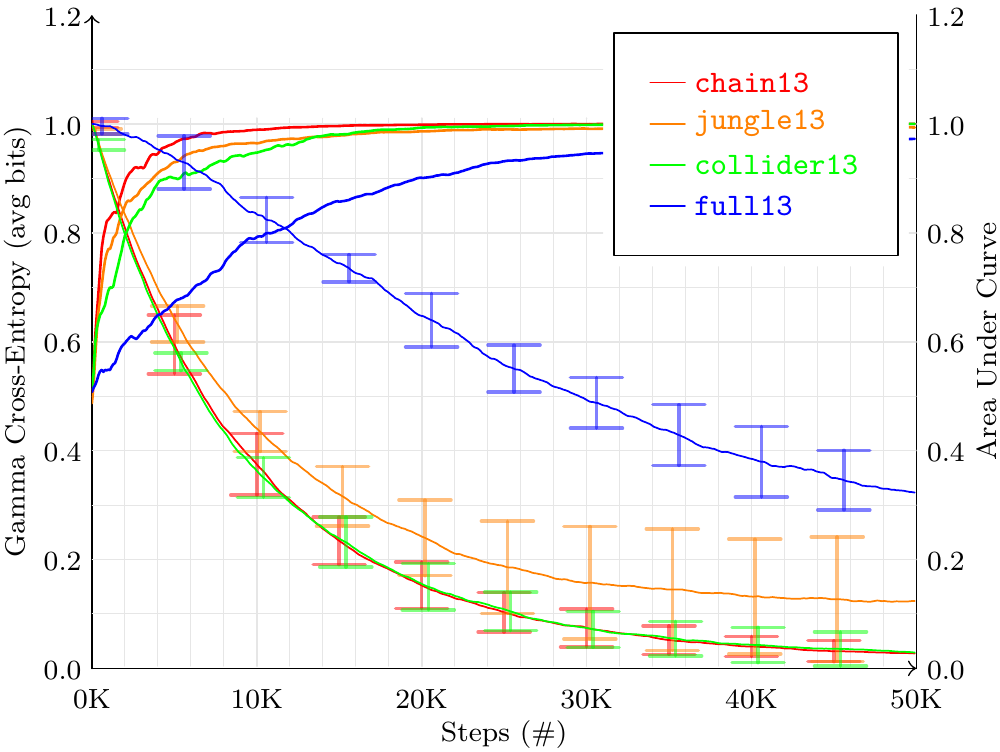}
     \caption{\textbf{Left to right, top to bottom} Average cross-entropy loss of edge beliefs $\sigmoid(\gamma)$ and Area-Under-Curve throughout training for the synthetic graphs \texttt{chainN}, \texttt{jungleN}, \texttt{colliderN} and \texttt{fullN}, $N$=3-13, grouped by graph size. Error bars represent $\pm 1\sigma$ over PRNG seeds 1-5.}\label{fig:sparsity}
\end{figure}

\setcounter{subsection}{12}

\subsection{Effect of regularization}\label{appendix_regularization}
\setcounter{topnumber}{8}
\setcounter{bottomnumber}{8}
\setcounter{totalnumber}{8}

\paragraph{Importance of sparsity regularizer.}
We use a $L1$ regularizer on the structure parameters $\gamma$ to encourage a sparse representation of edges in the causal graph. In order to better understand the effect of the $L1$ regularizer, we conducted ablation studies on the $L1$ regularizer. It seems that the regularizer has an small effect on rate of converges and that the model converges faster with the regularizer, This is shown in Figure \ref{fig::sparsity}. However, this does not seem to affect the final value the model converges to, this is shown in Table \ref{table:regularizer_comparison}.

\begin{figure}[t]
    \includegraphics[scale=0.25]{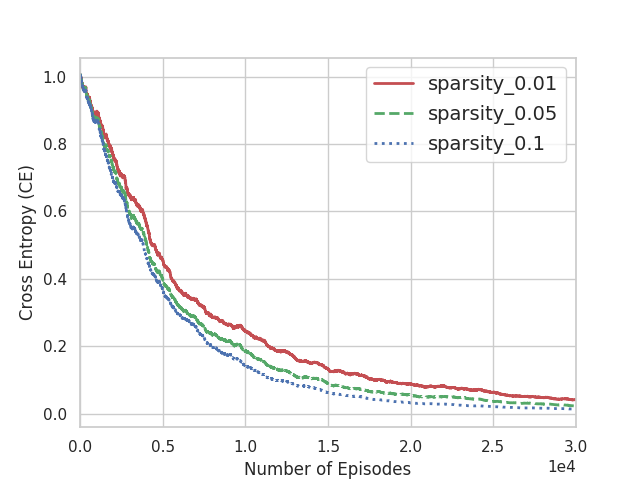}
    \includegraphics[scale=0.25]{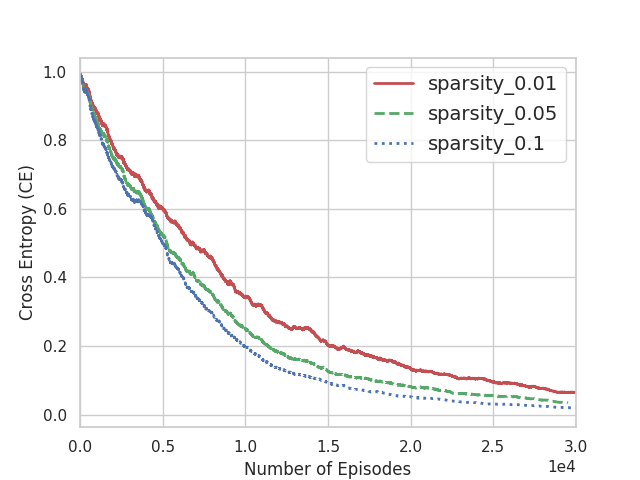}
    \includegraphics[scale=0.25]{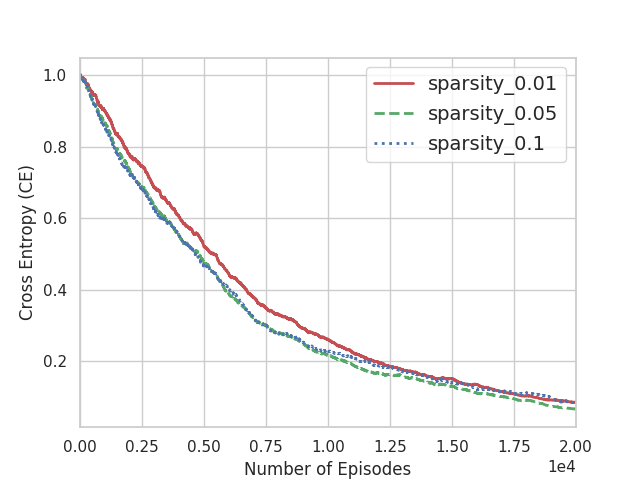}
    \caption{\textbf{Effect of sparsity (${lsparse}$) regularizer :} On 5 variable, 6 variable and 8 variable Nodes}
    \label{fig::sparsity}
\end{figure}

\begin{figure}
\centering
\begin{minipage}{.45\textwidth}
    \centering
    \includegraphics[scale=0.55]{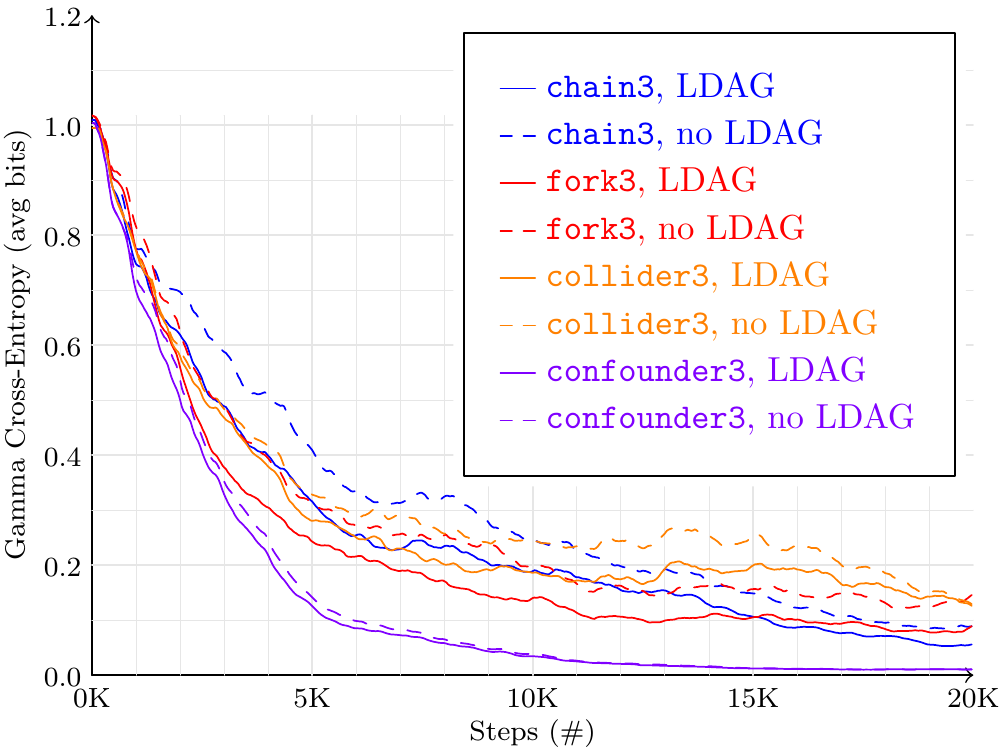}
    \caption{Ablations study results on all possible 3 variable graphs. Graphs show the cross-entropy loss on learned vs ground-truth edges over training time. Comparisons of model trained with and without DAG regularizer ($L_\textrm{DAG}$), showing that DAG regularizer helps convergence.}
    \label{fig:appendix_dag}
\end{minipage}
\;\;
\begin{minipage}{.45\textwidth}
    \centering
    \includegraphics[scale=0.55]{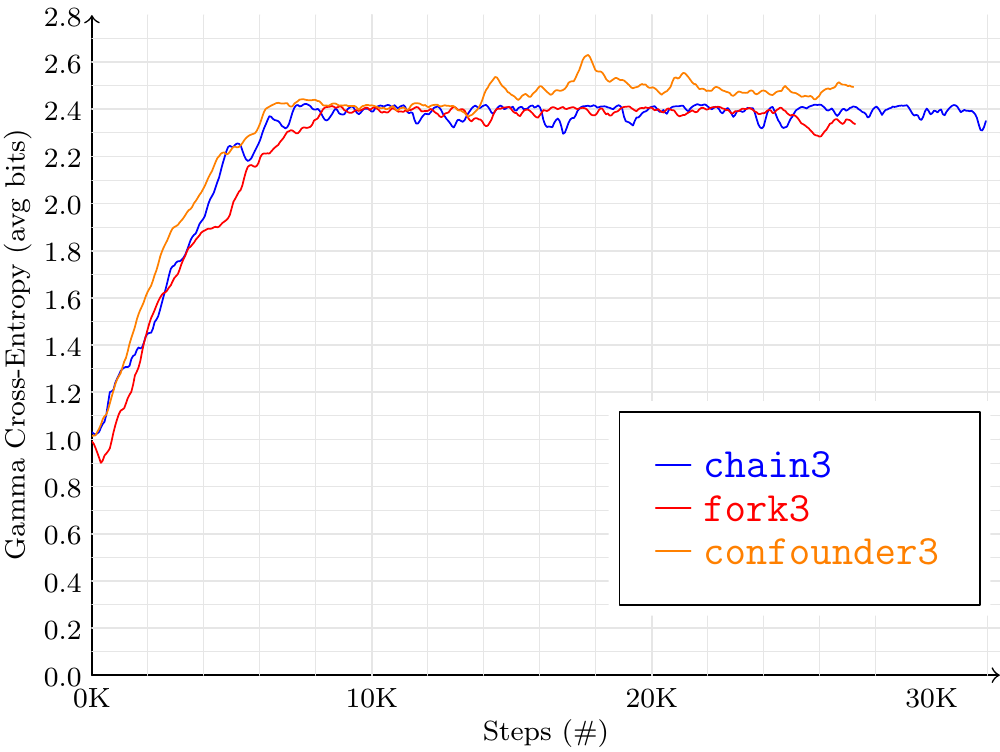}
    \caption{Edge CE loss for 3-variable graphs with no dropout when training functional parameters, showing the importance of this dropout.}
    \label{fig:gamma_4_7_CE_dropout}
\end{minipage}
\vspace{-1\baselineskip}
\end{figure}

\paragraph{Importance of DAG regularizer.} We use an acyclic regularizer to discourage length-2 cycles in the learned model. We found that for small models ($\le 5$ variables), the acyclic regularizer helps with faster convergence, without improving significantly the final cross-entropy. This is illustrated for the 3-variable graphs in Figure \ref{fig:appendix_dag}. However, for graphs larger than 5 variables, the acyclic regularizer starts playing an important role in encouraging the model to learn the correct structure. This is shown in the ablation study in Table \ref{table:regularizer_comparison}.

\subsection{Importance of dropout}\label{appendix_dropout}
To train the functional parameters on an observational distribution, one would need sampling adjacency matrices. One may be tempted to make these ``complete directed graph'' (all-ones except for a zero diagonal), to give the MLP maximum freedom to learn any potential causal relations itself. We demonstrate that functional parameter training cannot be carried out this way, and that it is necessary to ``drop out'' each edge (with probability of the current $\gamma$ value in our experiments) during pre-training of the conditional distributions of the SCM. We attempt to recover the previously-recoverable graphs \texttt{chain3}, \texttt{fork3} and \texttt{confounder3} without dropout, but fail to do so, as shown in Figure \ref{fig:gamma_4_7_CE_dropout}.

\begin{table}[b]
\vspace{-1\baselineskip}
\centering  

\label{table:baseline_ce}
{\begin{tabular}{lcccc}
\toprule
 & {\bf \gls{SDI}} & {\bf \citet{eaton2007exact}}  \\
\midrule
\texttt{Asia} & 0  & 0  \\
\texttt{chain8} & 0  & 0  \\
\texttt{jungle8} & 0  & 0  \\
\texttt{collider7} & 0  & 7  \\
\texttt{collider8} & 0.0  & 7  \\
\texttt{full8} & 0.0  & 1 \\
\bottomrule
\end{tabular} \\

\caption{\textbf{Comparisons:} Structured hamming distance (SHD) on learned and ground-truth edges on \texttt{asia} and various synthetic graphs. \citet{eaton2007exact} can not scale to larger variables graphs as shown in Table \ref{table:all_baseline_hamming}, hence, we compare to the largest graph that \citep{eaton2007exact} can scale up to. \gls{SDI} is compared to \citep{eaton2007exact} for \texttt{collider7}, \texttt{collider8} and \texttt{full8}, \citep{eaton2007belief} asserts with 100\% confidence a no-edge where there is one (false negative). For comparisons with all other methods \ref{table:all_baseline_hamming}.}
}
\end{table}


\begin{figure}[b]
\centering
\begin{minipage}{.3\textwidth}
    \centering
    \includegraphics[scale=0.3]{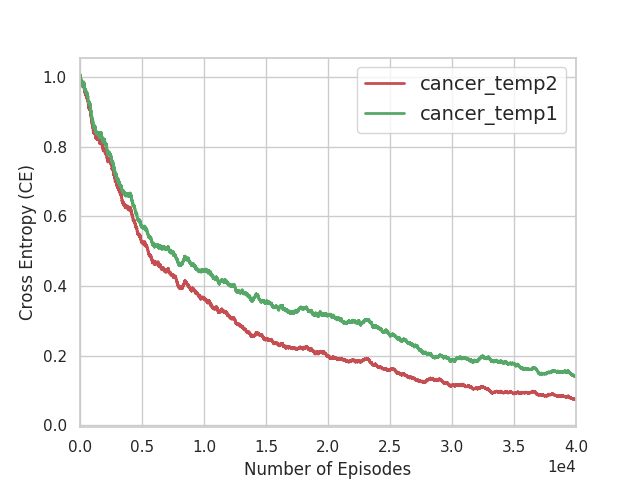}
    \caption{Cross-entropy for edge probability between learned and ground-truth SCM for Cancer at varying temperatures.}
    \label{fig:cancer_temp}
\end{minipage}
\;\;\;\;
\begin{minipage}{.6\textwidth}
    \centering
    \includegraphics[scale=0.2]{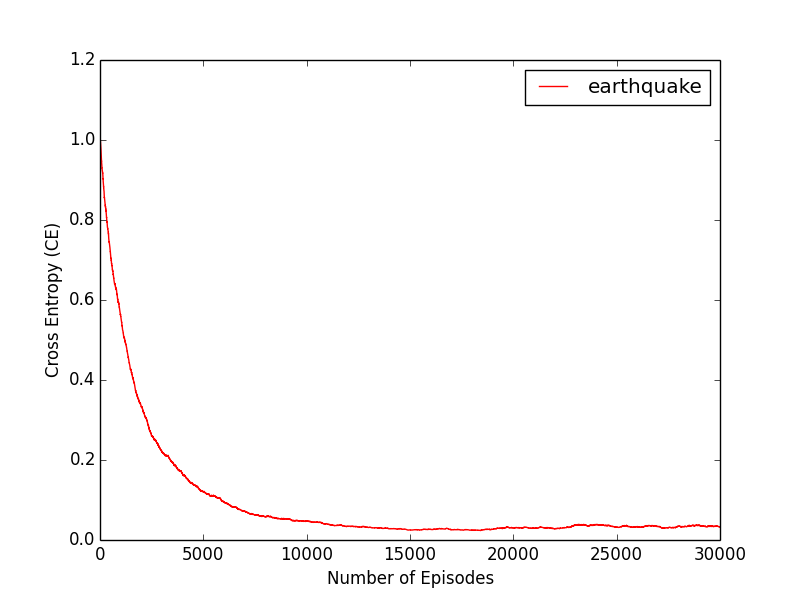}
    \includegraphics[scale=0.2]{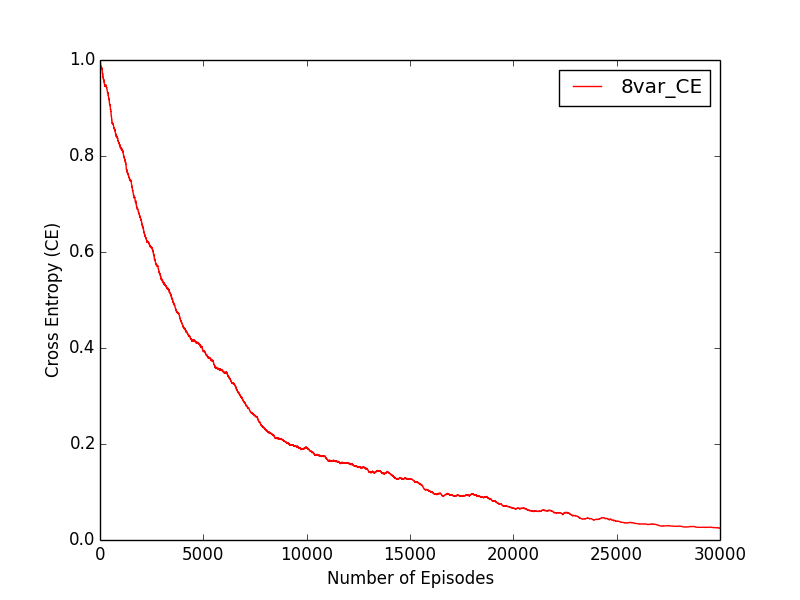}
    \caption{Cross-entropy for edge probability between learned and ground-truth SCM. \textbf{Left}: The Earthquake dataset with 6 variables. \textbf{Right}: The Asia dataset with 8 variables}
    \label{fig:earthquake_CE}
\end{minipage}
\end{figure}

\end{document}